\documentclass[preprint,12pt]{elsarticle}

\usepackage{booktabs}
\usepackage{multirow}

\usepackage{amssymb}
\usepackage{amsmath,amsfonts}
\usepackage{algorithmic}
\usepackage{array}
\usepackage{subfig}
\usepackage{textcomp}
\usepackage{stfloats}
\usepackage{url}
\usepackage{graphicx}
\usepackage{subcaption}
\usepackage[ruled]{algorithm2e}
\usepackage{tabularray}
\usepackage{color}
\usepackage{wrapfig}

\journal{Information Fusion}

\begin{document}

\begin{frontmatter}

\title{FDiff-Fusion: Denoising diffusion fusion network based on fuzzy learning for 3D medical image segmentation}

\author{Weiping~Ding\corref{cor1}}
\ead{dwp9988@163.com}
\author{Sheng~Geng}
\author{Haipeng~Wang}
\author{Jiashuang~Huang}
\author{Tianyi~Zhou}

\address{School of Artificial Intelligence and Computer Science, Nantong University,
            Nantong,
            226019,
            China}




\cortext[cor1]{Corresponding author}

\begin{abstract}
In recent years, the denoising diffusion model has achieved remarkable success in image segmentation modeling. With its powerful nonlinear modeling capabilities and superior generalization performance, denoising diffusion models have gradually been applied to medical image segmentation tasks, bringing new perspectives and methods to this field. However, existing methods overlook the uncertainty of segmentation boundaries and the fuzziness of regions, resulting in the instability and inaccuracy of the segmentation results. To solve this problem, a denoising diffusion fusion network based on fuzzy learning for 3D medical image segmentation (FDiff-Fusion) is proposed in this paper. By integrating the denoising diffusion model into the classical U-Net network, this model can effectively extract rich semantic information from input medical images, thus providing excellent pixel-level representation for medical image segmentation. In this paper, a fuzzy learning module is designed on the skip path of U-Net network because of the widespread boundary uncertainty and region blurring of medical image segmentation. The module sets several fuzzy membership functions for the input encoded features to describe the similarity degree between the feature points, and applies fuzzy rules to the fuzzy membership functions, thus enhancing the modeling ability of the model for uncertain boundaries and fuzzy regions. In addition, in order to improve the accuracy and robustness of the model segmentation results, we introduced an iterative attention feature fusion method in the test phase, which added local context information to the global context information in the attention module to fuse the prediction results of each denoising time step. Finally, to validate the effectiveness of FDiff-Fusion, we compare it with existing advanced segmentation networks on the BRATS 2020 brain tumor dataset and the BTCV abdominal multi-organ dataset. The results show that FDiff-Fusion significantly improves the Dice scores and HD95 distance on these two datasets, demonstrating its superiority in medical image segmentation tasks.
\end{abstract}


\end{frontmatter}


\section{Introduction}
\label{Introduction}
Brain tumors refer to abnormal cell proliferation within brain tissue or its associated structures, which can be either benign or malignant. Benign brain tumors typically grow slowly and do not easily spread, but due to their specific location, they can still significantly impact brain function. Malignant brain tumors, on the other hand, are characterized by rapid growth and invasion of surrounding tissues, generally leading to a poor prognosis and posing a significant threat to the patient's life and health. According to statistics, hundreds of thousands of people around the world are diagnosed with brain tumors every year, and the incidence of malignant tumors is on the rise \cite{r1}. At present, the early diagnosis and treatment of brain tumors have become one of the focuses of clinical attention, however, due to their complex location and structure within the brain tissue, as well as the heterogeneity of different types of tumors, making their diagnosis and treatment extremely challenging. In addition, segmentation of multiple organs in the abdomen is an important medical image processing task, aiming to accurately extract the contours and boundaries of multiple organs, such as liver, kidney, pancreas, etc., from abdominal images to help doctors in diagnosis, surgical planning and treatment monitoring \cite{r2}. Traditional methods used for brain tumor and abdominal multi-organ segmentation often require doctors to possess rich experience and professional knowledge. Moreover, the segmentation process is complex and time-consuming, making it susceptible to subjective factors and errors. Therefore, automated medical image segmentation techniques play a crucial role in enhancing diagnostic accuracy and clinical efficiency. They aid physicians in promptly identifying lesion areas, thereby improving patient treatment outcomes and survival rates.

Medical image segmentation is a process that accurately delineates or separates structures or regions of interest in medical images using computer vision and image processing techniques \cite{r3}. It plays a vital  role in medical image analysis, aiding physicians in diagnosis, disease monitoring, and treatment planning. By automatically identifying and quantifying lesion areas in images, physicians can devise treatment plans more rapidly and precisely, thereby improving patient outcomes. However, due to the complexity and diversity of medical images, such as noise, segmentation edge uncertainty and target region blurring, medical image segmentation still faces numerous challenges and difficulties \cite{r4}.

With the continuous advancement of deep learning, Convolutional Neural Networks (CNNs) play a key role in medical image segmentation due to their powerful feature representation capabilities \cite{r5,r6}. In recent years, the performance of medical image segmentation has been further improved by introducing Transformer structures. Research indicates that CNNs and Transformers effectively learn more representative features in medical images, leading to significant performance improvements in various segmentation tasks \cite{r7,r8,r9}. Among these, Ronneberger et al. \cite{r10} proposed U-Net, which features a unique U-shaped architecture comprising symmetrical encoder and decoder paths, connected through skip connections. This architectural design helps capture both local and global information within images, thereby enhancing segmentation accuracy. UNETR \cite{r11} utilizes a Vision Transformer (ViT) as the encoder to directly model global features, and employs a decoder with skip connections from a CNN to output segmentation results. SwinUNETR leverages the Swin-Transformer as an encoder to extract multi-scale features and employs a CNN-based decoder to generate output results, achieving state-of-the-art performance in medical image segmentation \cite{r12,r13}. Pang et al.\cite{r14} proposed Slim UNETR, which achieves information exchange in self-attention mechanism decomposition and cost-effective representation aggregation through the Slim UNETR Block. This model enables effective segmentation on resource-constrained devices. Wang et al. \cite{r15} proposed TransBTS, which initially uses 3D CNNs to extract volumetric spatial features for capturing local 3D contextual information. Each volume is then reshaped into tokens and fed into a Transformer for global feature modeling. Finally, the decoder leverages the Transformer's feature embeddings and performs progressive upsampling to predict the full-resolution segmentation map. Lin et al. \cite{r16} proposed CKD-TransBTS, which extracts multimodal features using a dual-branch hybrid encoder and modality-correlated cross-attention (MCCA) block. In the decoder, they designed the Trans\&CNN Feature Calibration (TCFC) block to fuse features from both Transformer and CNN, demonstrating promising results in segmentation on the BraTS 2021 dataset. Chen et al. \cite{r17} proposed TransUNet, which combines the strengths of Transformers and U-Net. The Transformer encodes tokenized image patches from feature maps extracted by CNN to capture global contextual information. Subsequently, the decoder upsamples the encoded features and combines them with high-resolution CNN feature maps to achieve precise localization. Yang et al. \cite{r18} proposed EPT-Net, which enhances 3D spatial localization capabilities through dual position transformers and introduces an edge-weight guidance module to extract edge information from low-level features. This design leverages rich fine-grained information to improve segmentation accuracy without increasing network parameters.

In the field of medical image processing, denoising has always been a fundamental and crucial issue. Traditional denoising methods typically rely on filters or transform domain processing, such as wavelet transforms, mean filtering, and Gaussian filtering. However, these methods have limited effectiveness in handling complex noise and preserving image details \cite{r19,r20}. In recent years, with the rapid development of deep learning technology, the denoising diffusion model has gradually received widespread attention. Its basic principle is to use a deep neural network model, usually a U-Net network model, to map the input image containing noise to the corresponding clean image space through training \cite{r21,r22}. Compared with traditional medical image segmentation methods, the denoising diffusion model has stronger nonlinear construction model ability and superior generalization performance, which can more effectively capture complex structures and detailed information in images. In MedSegDiff proposed by Wu et al. \cite{r23}, in order to enhance the step-by-step regional attention in diffusion probabilistic models (DPM) for medical image segmentation, they proposed dynamic conditional coding, which establishes state-adaptive conditions for each sampling step. And, they further proposed a feature frequency resolver to eliminate the negative impact of high-frequency noise components in the process. Building upon MedSegDiff, Wu et al. propose a novel Transformer-based conditional U-Net framework and a new spectrum space Transformer to model the interaction between noise and semantic features. This improved architecture gives rise to a new diffusion-based medical image segmentation method called MedSegDiff-V2 \cite{r24}, significantly enhancing the segmentation performance of MedSegDiff. Xing et al. \cite{r25} proposed Diff-UNet for medical volume segmentation. They integrate the diffusion model into a standard U-shaped architecture, effectively extracting semantic information from input volumes, thereby providing excellent pixel-level representations for medical volume segmentation. To enhance the robustness of the diffusion model's prediction results, they also introduce a fusion module based on stepwise uncertainty during inference to merge the outputs of diffusion models at each step.Wolleb et al. \cite{r26} utilize diffusion models to address the problem of two-dimensional medical image segmentation, and during testing, they enhance segmentation robustness by summing the output results of each diffusion step for fusion.

Despite the significant success of denoising diffusion models in the field of image segmentation, there are still some challenges and problems to be solved for medical image segmentation. For example, medical images have large individual differences and are often subject to various forms of noise interference, includes signal interference from electronic devices, motion artifacts during imaging, and noise based on physical properties.  Moreover, the segmentation boundary uncertainty and region blurring phenomenon of the segmentation target in medical images are common. The above models struggle to effectively handle uncertain segmentation boundaries and blurry regions in medical images.

To address these problems, we propose a denoising diffusion fusion network based on fuzzy learning for 3D medical image segmentation, named FDiff-Fusion. This segmentation method enhances the modeling capability of U-Net networks for uncertain segmentation boundaries and blurry regions in medical images by introducing a fuzzy learning module (FLM). Additionally, a fusion module based on iterative attention mechanism (IAF) is designed to fuse the prediction results of multiple time steps in the denoising process, enabling more accurate segmentation results during testing. The proposed approach aims to improve the accuracy and robustness of medical image segmentation, and to deal with the cases of boundary uncertainty and region ambiguity more effectively.  This is crucial for enhancing medical image segmentation techniques and improving the accuracy of clinical diagnosis.

The main contributions of this paper are as follows.
\begin{itemize}
    \item {To address the uncertainty of segmentation boundaries and the fuzziness of regions in medical image segmentation, we introduced a fuzzy learning module. This module sets multiple fuzzy membership functions for input encoding features to describe the similarity between feature points. By applying fuzzy rules to these fuzzy membership functions, it enhances the model's capability to model the uncertainty of segmentation boundaries and the fuzziness of regions.}
    \item {To enhance the accuracy and robustness of the model's prediction results, we propose a fusion module based on an iterative attention mechanism. This module optimizes the final segmentation results by incorporating the local contextual information of each time step segmentation image into the global contextual information of the attention module.}
    \item {To comprehensively evaluate the segmentation performance of the FDiff-Fusion, we conduct comparative experiments using multiple state-of-the-art deep learning models on the BRATS 2020 brain tumor dataset and BTCV abdominal multi-organ dataset. The experimental results demonstrate the effectiveness of our model.}
\end{itemize}

The rest of this paper is organized as follows. Section \ref{basic} introduces the relevant background knowledge, mainly focusing on the U-Net network model, fuzzy learning theory, and denoising diffusion models.  Section \ref{FDiff-Fusion} presents the architecture of the proposed FDiff-Fusion model.  Section \ref{result} discusses the experimental results and analyzes them. Section \ref{conclusion} provides a summary and outlook.

\section{PRELIMINARIES\label{basic}}

In this section, we provide a brief overview of the foundational knowledge discussed in this paper.  Section \ref{U-Net} introduces the U-Net network model. Section \ref{Fuzzy learning theory} presents fuzzy learning theory. Section \ref{Denoising Diffusion Model} discusses denoising diffusion models.

\subsection{U-Net network model\label{U-Net}}

The core idea of U-Net network is to combine encoder and decoder to achieve end to end image segmentation. Its unique U-shaped structure enables the network to capture both global and local information of the image simultaneously, effectively overcoming some limitations of traditional methods. Additionally, U-Net utilizes skip connections, allowing the network to transmit information between different levels, further enhancing segmentation accuracy and robustness. The network structure comprises:

\noindent \textbf{Encoder (contracting path):} The encoder downsamples the input image multiple times to extract high-level feature representations. Typically constructed using convolutional and pooling layers, the encoder gradually reduces the spatial resolution of the feature maps while increasing the number of channels, capturing richer feature information. This process helps refine detailed information in the input image into higher-level abstract features, which provides the basis for the subsequent image segmentation task.

\noindent \textbf{Decoder (expansive path):} The decoder upsamples the feature maps extracted by the encoder and concatenates them with corresponding encoder layer feature maps to restore the spatial resolution of the original image and generate segmentation results. The decoder includes deconvolutional layers and skip connections, with skip connections aiding the network in more fully utilizing feature information from different levels to improve segmentation accuracy.

\noindent \textbf{Skip connections:} Skip connections in the U-Net network are one of its key features. Through skip connections, feature maps from the encoder can be directly connected to corresponding feature maps in the decoder. This connection mechanism allows the network to more fully utilize feature information from different levels, preventing information loss and thereby improving segmentation accuracy and robustness.

\subsection{Fuzzy learning theory\label{Fuzzy learning theory}}
Fuzzy learning is a machine learning approach based on fuzzy set theory, aimed at handling the fuzziness or uncertainty in data \cite{r27}. Fuzzy set theory is used to describe the fuzzy relationship between things and to model the uncertainty. Here are some key concepts and methods of fuzzy learning:

\noindent\textbf{Fuzzy Set Theory:} Fuzzy set theory introduces the concept of fuzziness to describe uncertainty and vagueness between things. In this theory, an element can partially belong to a set, rather than completely belonging or not belonging, and this is mainly indicated by membership functions \cite{r28,r29}.

\noindent\textbf{Fuzzy Inference:} Fuzzy inference deals with fuzzy problems by introducing fuzzy sets and fuzzy logical operations. A fuzzy set is a function of membership degree of an element between 0 and 1, representing the degree to which the element belongs to the set. Fuzzy logic operations, including fuzzy and, fuzzy or, fuzzy not operations, are used to deal with the relationship between fuzzy propositions. The basic idea is to infer based on fuzzy rules, where fuzzy rules consist of a condition part and a conclusion part. The condition part is a set of fuzzy propositions, and the conclusion part is the corresponding fuzzy conclusion. Fuzzy inference introduces a fuzzy inference engine, which infers based on fuzzy rules and input fuzzy propositions to obtain fuzzy conclusions \cite{r30,r31}.

\noindent\textbf{Edge Detection:} Fuzzy learning can also be utilized for edge detection tasks in medical images. Traditional medical image segmentation algorithms often struggle to effectively handle blurry boundaries or complex background situations, while fuzzy learning methods can better identify fuzzy edges and improve the accuracy of edge detection \cite{r32,r33}.

\noindent\textbf{Fuzzy Membership Functions:} In some traditional image segmentation algorithms, fuzzy membership functions can be introduced to describe the membership relationship between pixels and different tissues or structures. These fuzzy membership functions can be learned and optimized through fuzzy learning methods, thus more accurately characterizing the fuzziness and uncertainty in the image \cite{r34,r35}.

\subsection{Denoising Diffusion Model\label{Denoising Diffusion Model}}

\begin{figure*}[!ht]
\centering
\includegraphics[width=\textwidth]{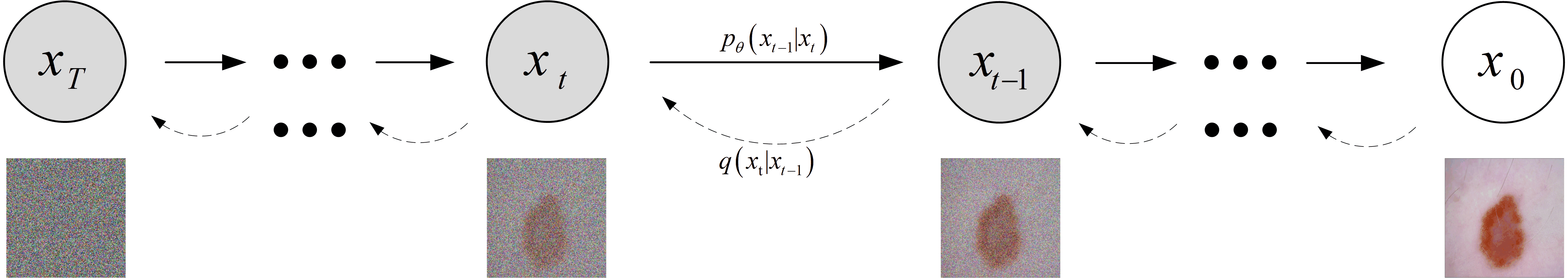}
\caption{Forward process and reverse process of denoising diffusion model. The dashed arrows represent the forward process of the denoising diffusion model, and the solid arrows represent the reverse process of the denoising diffusion model.}
\label{fig1}
\end{figure*}

The denoising diffusion model consists of a forward process and a reverse process. In the forward process, noise is gradually applied to the image until it is corrupted into a completely Gaussian-noised image. Then, in the reverse process, a deep neural network model is used to learn the process of restoring the original image from the Gaussian-noised image \cite{r36,r37}. Specifically, as shown in Figure \ref{fig1}, the solid lines represent the reverse process, while the dashed lines represent the forward process. In the forward process, Gaussian noise is gradually added to the original image $x_0$, and each step of the resulting image $x_t$ is only related to the image obtained in the previous step $x_{t-1}$ until the image $x_T$ of the $T$ step becomes a pure Gaussian noise image, which can be regarded as a Markov process. The reverse process is then the gradual removal of Gaussian noise, starting with the given Gaussian-noised image $x_T$, and proceeding step by step until it is restored to the original image $x_0$. Specifically, in the forward process, for the original image $x_0$, variational inference is performed over $T$ time steps of the Markov process to learn the training data distribution. This process can be represented as:
\begin{equation}
\label{1}
q(x_{t}|x_{t-1})=\mathcal{N}(x_{t};\sqrt{\alpha_{t}}x_{t-1},(1-\alpha_{t})I)
\end{equation}

\noindent where $t$ represents the $t$-th time step in the Markov chain. $\alpha_t=1-\beta_t$ is the noise controller that controls the strength of noise added in the $t$th time step, and $\alpha_{t}\in(0,1)$ and $\alpha_{1}>\cdots>\alpha_{T}$.
In the reverse process, a deep neural network, typically a U-Net network, is trained to learn to remove the added Gaussian noise, resulting in the restoration of the pure noise image $x_T$ to the original image $x_0$. This process can be represented as:
\begin{equation}
\label{2}
p_\theta(x_{0:T-1}|x_T)=\prod_{t=1}^Tp_\theta(x_{t-1}|x_t)
\end{equation}

\noindent where $p_{\theta}(\cdot)$ is the deep learning model and $\theta$ is its parameter. The reverse process primarily aims to obtain clear segmentation results by removing the added Gaussian noise.

\begin{figure*}[!ht]
\centering
\includegraphics[width=\textwidth]{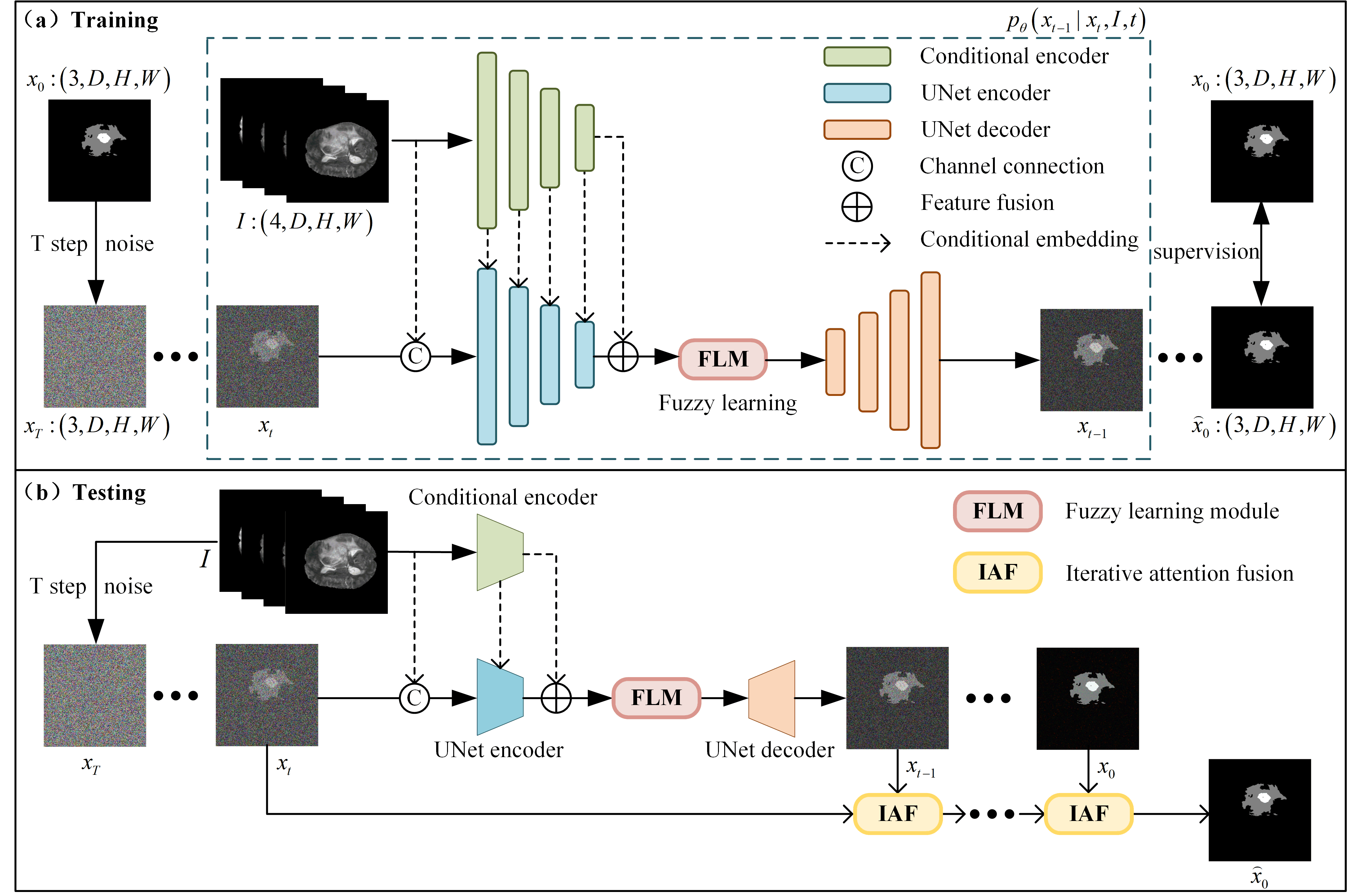}
\caption{Framework of our proposed FDiff-Fusion.(a) is the training phase of FDiff-Fusion, which learns the denoising process by adding U-Net with fuzzy learning module and CNN for conditional information embedding. (b) is the testing phase of FDiff-Fusion, which gradually denoises the noisy image and fuses the prediction graph generated by each time step by iterating attention module to get the final segmentation graph.}
\label{fig2}
\end{figure*}

\section{FDiff-Fusion\label{FDiff-Fusion}}
Traditional medical image segmentation methods typically involve directly using the medical images as input and training models to predict corresponding segmentation label maps \cite{r38}. However, unlike this direct input approach, the denoising diffusion models adopt a different strategy. The learning process of the denoising diffusion models does not directly predict segmentation label maps but focuses on learning the process of removing Gaussian noise \cite{r39,r40}. Specifically, during the training phase, the diffusion model takes medical images along with segmentation label maps with Gaussian noise as input and then learns to remove Gaussian noise from the medical images using a deep learning model to generate clear segmentation results. The key to this method lies in improving the quality of segmentation results through the denoising process rather than directly segmentation and prediction of medical images \cite{r41,r42}. However, despite the decent achievements of denoising diffusion models in the field of image segmentation, there are still challenges in the processing of segmentation boundary uncertainty and regional blurring in medical images. To overcome these challenges and enhance the accuracy and robustness of segmentation results, we proposes a denoising diffusion fusion network based on fuzzy learning for 3D medical image segmentation, named FDiff-Fusion, as shown in Figure \ref{fig2}.

In this section, the denoising diffusion fusion network based on fuzzy learning for 3D medical image segmentation is divided into three parts. Firstly, Section \ref{3.1} introduces the FDiff-Fusion model; Section \ref{3.2} presents the fuzzy learning module; and Section \ref{3.3} describes the iterative attention fusion. In addition, we also provide a symbol table to explain the symbols involved in this article, as shown in Table \ref{table0}.

\begin{table}
\caption{Description of notations}
\label{table0}
\centering
\resizebox{\linewidth}{!}{%
\begin{tabular}{cl} 
\hline
Notation & Description  \\ 
\hline
$q(\cdot)$         & The forward process              \\
$p(\cdot)$         & The reverse process             \\
$\mathcal{N}(\cdot)$         &The normal distribution              \\
$t$         &The t-th time step, t=1,2,...,T               \\
$\beta_{t}$         & A hyperparameter sequence that controls the strength of the noise            \\
$\epsilon_{t}$         &Noise randomly sampled from a standard normal distribution              \\
$p_{\theta}(\cdot)$         &Deep learning model for learning the denoising process              \\
$FLM$         & Fuzzy learning module             \\
$IAF$         & Iterative attention fusion    \\
\hline
\end{tabular}
}
\end{table}

\subsection{FDiff-Fusion model}
\label{3.1}
The FDiff-Fusion model consists of two processes, namely the forward process and the reverse process. In the forward process, the original medical image $x_{0}\in\mathbb{R}^{C\times D\times H\times W}$  is progressively contaminated with Gaussian noise to obtain a pure noisy image $x_{T}\in\mathbb{R}^{C\times D\times H\times W}$. Since the image $x_t$  in the forward process only depends on the previous step $x_{t-1}$, this process can be considered as a Markov process, satisfying:

\begin{equation}
\label{3}
q(x_t\mid x_{t-1})=\mathcal{N}\big(x_t;\sqrt{1-\beta_t}x_{t-1},\beta_t\mathbf{I}\big)
\end{equation}

\begin{equation}
\label{4}
q(x_{1:T}\mid x_{0})=\prod_{t=1}^{T}q(x_{t}\mid x_{t-1})
\end{equation}

\noindent where $q(\cdot)$ represents the forward process, $\mathcal{N}(\cdot)$ denotes the normal distribution, and $t$ indicates the $t$-th time step. $\beta_{t}\in(0,1)$ is a hyperparameter sequence controlling the noise level added at each time step. This sequence is a pre-defined fixed coefficient, satisfying the condition $\beta_{1}<\beta_{2}<\cdots<\beta_{T}$, and it linearly increases from 0.0001 to 0.02.

Subsequently, using the reparameter technique, we sample $x_t$ through Equation (\ref{3}), that is, generate a random variable $\epsilon_{t}\sim\mathcal{N}(0,1)$, so that $\alpha_{t}=1-\beta_{t}$, which can be obtained:

\begin{equation}
\label{5}
x_t=\sqrt{\bar{\alpha}_t}x_0+\sqrt{1-\bar{\alpha}_t}{\epsilon}_t
\end{equation}

\noindent where $\bar{\alpha}_{t}=\prod_{i=1}^{T}\alpha_{t}$. In the reverse process, since we cannot directly infer $q(x_{t-1}\mid x_{t})$, we use a deep learning model (U-Net) to fit the distribution $q(x_{t-1}\mid x_{t})$, with model parameters $\theta $:
\begin{equation}
\label{6}
p_\theta(x_{0:T})=p(x_T)\prod_{t=1}^Tp_\theta(x_{t-1}|x_t)
\end{equation}
\begin{equation}
\label{7}
p_\theta(x_{t-1}\mid x_t)=\mathcal{N}(x_{t-1};\mu_\theta(x_t,t),\Sigma_\theta(x_t,t))
\end{equation}

\noindent where $p_{\theta}(\cdot)$ is the deep learning model used to fit the distribution $q(x_{t-1}\mid x_{t})$. $\mu_{\theta}$ represents the mean and $\Sigma_{\theta}$ represents the variance, where $t$ is the denoising time step. Although $q(x_{t-1}\mid x_{t})$ cannot be directly derived, knowing $x_0$, we can obtain $q(x_{t-1}\mid x_{t},x_{0})$ through Bayesian formula:

\begin{equation}
\label{8}
q(x_{t-1}\mid x_{t},x_{0})=\mathcal{N}(x_{t-1};\tilde{\mu}(x_{t},x_{0}),\tilde{\beta}_{t}\mathrm{I})
\end{equation}
\begin{equation}
\label{9}
q(x_{t-1}|x_t,x_0)=\frac{q(x_t|x_{t-1},x_0)q(x_{t-1}|x_0)}{q(x_t|x_0)}
\end{equation}
\noindent where $\tilde{\mu}(x_t,x_0)$ and $\tilde{\beta}_{t}$ represent the true values of the mean and variance of the reverse process, respectively.

From Equation (\ref{9}), it can be seen that the reverse process of the denoising diffusion model can be converted into a forward process by Bayesian formula. And get the probability density function and the gaussian probability density function of the index part
$exp\left(-\frac{(x-\mu)^{2}}{2\sigma^{2}}\right)=exp\left(-\frac{1}{2}\left(\frac{1}{\sigma^{2}}x^{2}-\frac{2\mu}{\sigma^{2}}x+\frac{\mu^{2}}{\sigma^{2}}\right)\right)$ to corresponding, namely:
\begin{equation}
\label{10}
\tilde{\beta}_t=\frac{1-\bar{\alpha}_{t-1}}{1-\bar{\alpha}_t}\cdot\beta_t
\end{equation}
\begin{equation}
\label{11}
\tilde{\mu}_{t}(x_{t},x_{0})=\frac{\sqrt{\alpha_{t}}(1-\bar{\alpha}_{t-1})}{1-\bar{\alpha}_{t}}x_{t}+\frac{\sqrt{\bar{\alpha}_{t-1}}\beta_{t}}{1-\bar{\alpha}_{t}}x_{0}
\end{equation}

According to Equation (\ref{5}), it can be obtained:
\begin{equation}
\label{12}
x_0=\frac{1}{\sqrt{\bar{\alpha}_t}}(x_t-\sqrt{1-\bar{\alpha}_t}\epsilon_t)
\end{equation}

Bringing this into Equation (\ref{11}), we can get:
\begin{equation}
\label{13}
\tilde{\mu}_t=\frac{1}{\sqrt{\alpha_t}}(x_t-\frac{1-\alpha_t}{\sqrt{1-\bar{\alpha}_t}}\epsilon_t)
\end{equation}

Since we need to use the deep learning model $p_{\theta}$ to approximate the distribution $q(x_{t-1}\mid x_{t})$, according to Equation (\ref{10}), we train the model $\mu_\theta(x_t,t)$ to estimate Equation (\ref{13}). As $x_{t}$ is known during the training phase and is used as an input, we can instead let the model estimate the noise $\epsilon_{t}$, by setting:
\begin{equation}
\label{14}
\mu_{\theta}(x_{t},t)=\frac{1}{\sqrt{\alpha_{t}}}(x_{t}-\frac{1-\alpha_{t}}{\sqrt{1-\bar{\alpha}_t}}\epsilon_{\theta}(x_{t},t))
\end{equation}

According to Equation (\ref{14}), the final result is:
\begin{equation}
\label{15}
x_{t-1}=\mathcal N(x_{t-1};\frac{1}{\sqrt{\alpha_{t}}}(x_{t}-\frac{1-\alpha_{t}}{\sqrt{1-\bar{\alpha}_{t}}}\epsilon_{\theta}(x_{t},t)),\Sigma_{\theta}(x_{t},t))
\end{equation}

As shown in Figure \ref{fig2}, during the testing phase, a 3D medical image $I\in\mathbb{R}^{D\times H\times W\times C}$ is provided as conditional embedding for each time step of the denoising diffusion model, where $D$, $H$, $W$ and $C$ represent the depth, height, width, and channels of the 3D medical image, respectively. Firstly, $T$ steps of Gaussian noise are added to the 3D medical image $I$ to obtain a pure noisy image $x_T$. Then, a step-by-step denoising operation is performed on the pure noisy image $x_T$. Taking the $t$-th time step as an example, the noisy image $x_t$ and the original medical image $I$ are concatenated along the channels and fed into the encoder of a U-Net network to obtain multi-scale features $f_{t}\in\mathbb{R}^{C\times D\times H\times W}$. Simultaneously, to better incorporate the original 3D medical image as conditional embedding, $I$ is input into a deep convolutional neural network (CNN) to obtain multi-scale conditional features $f_{I}\in\mathbb{R}^{C\times D\times H\times W}$. Since $f_t$ and $f_I$ contain the same number and size of features, the corresponding scale features are added together to obtain fused features. However, due to the presence of uncertainty in the boundaries of segmentation targets and the blurring of regions in 3D medical images, simply adding the multi-scale features extracted by the encoder of the U-Net network and those extracted by the CNN as encoding features for each time step will struggle to address the problem of uncertain boundaries and blurry regions in segmentation targets. To address this issue, we propose a Fuzzy Learning Module (FLM).On the one hand, the original 3D medical image contains accurate segmentation targets, but it is difficult to determine their segmentation boundaries. On the other hand, the segmentation map at the current time step contains enhanced segmentation targets, but lacks accuracy. Therefore, we utilize a Fuzzy Learning Module (FLM) on the connection paths between the encoder and decoder of the U-Net network to process the encoding features with fuzzy feature treatment. Subsequently, the multi-scale features processed by the U-Net encoder are input into the decoder via skip connections for processing, resulting in the prediction result $x_{t-1}$ for the $t$-th time step. Finally, to improve the accuracy and robustness of the model segmentation, the segmentation maps obtained at each time step are fused through iterative attention to obtain the final segmentation result $\hat{x}_{0}\in\mathbb{R}^{C\times D\times H\times W}$.

In contrast to the testing phase where only the 3D medical image serves as input to the model, in the training phase, this paper simultaneously inputs the 3D medical image $I\in\mathbb{R}^{D\times H\times W\times C}$ and its corresponding segmentation label $x_{0}\in\mathbb{R}^{C\times D\times H\times W}$  into the model. Here, the 3D medical image acts as conditional information embedding, while the corresponding segmentation label is used to add Gaussian noise for $T$ steps to obtain the Gaussian noise image $x_{T}\in\mathbb{R}^{C\times D\times H\times W}$. After T steps of denoising operations, the model predicts the segmentation label image $\hat{x}_{0}\in\mathbb{R}^{C\times D\times H\times W}$, which is then used for loss calculation along with the segmentation label image. FDiff-Fusion is trained by combining the Mean Squared Error (MSE) Loss, Binary Cross-Entropy (BCE) Loss, and Dice Loss commonly used in traditional image segmentation networks. Therefore, the total loss $\mathcal{L}_{total}$ of the model is defined as:
\begin{equation}
\label{16}
\mathcal{L}_{total}=\mathcal{L}_{mse}(\hat{x}_{0},x_{0})+\mathcal{L}_{bce}(\hat{x}_{0},x_{0})+\mathcal{L}_{dice}(\hat{x}_{0},x_{0})
\end{equation}

\subsection{Fuzzy learning module}
\label{3.2}
\begin{figure}[!ht]
\centering
\includegraphics[width=\textwidth]{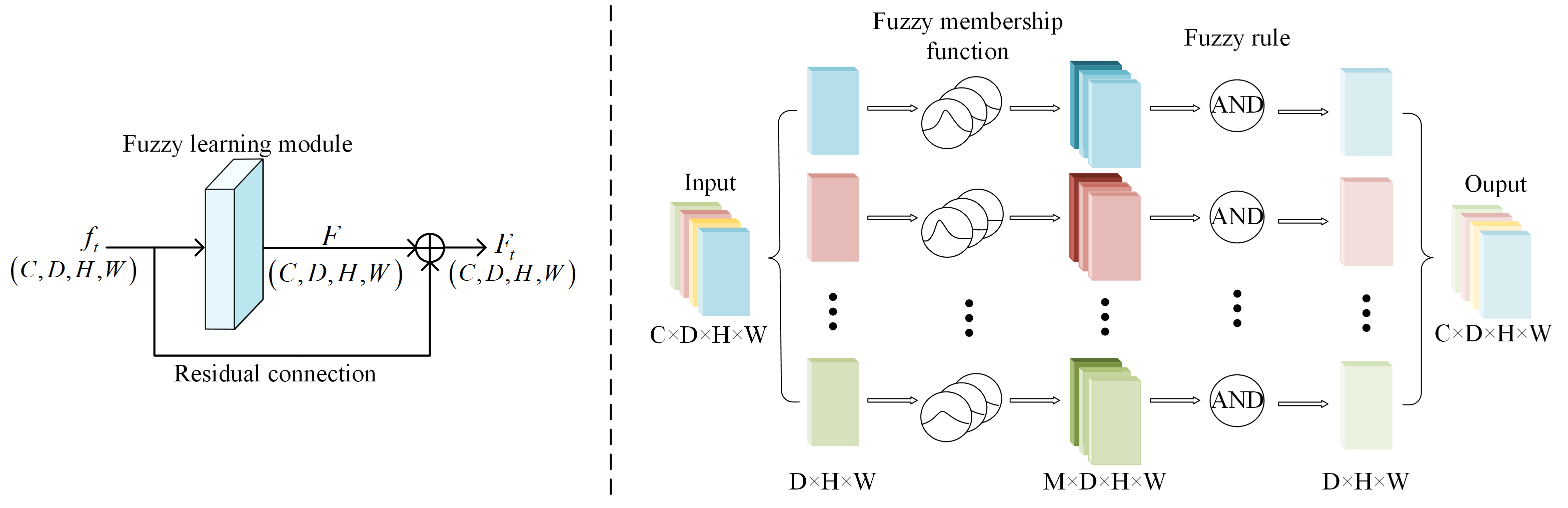}
\caption{Fuzzy Learning Module (FLM) architecture. The left part is a fuzzy residual connection, which fuses the encoding features with the fuzzy features through residual connections. The right part is fuzzy learning, which uses gaussian functions to calculate the membership degree of the encoded feature.}
\label{fig3}
\end{figure}

To solve the boundary uncertainty and region blurring in 3D medical images, we proposes a fuzzy learning module (FLM) to tackle this issue. Specifically, for the features extracted by the U-Net network, a fuzzy residual connection is designed along the skip pathways of the U-Net network for processing. The fuzzy residual connection is divided into a fuzzy learning module and a residual connection, where the fuzzy learning module comprises fuzzy membership functions and fuzzy rules. The goal of FLM is to model the complex rules between feature maps and the semantic category of each pixel. As shown in Figure \ref{fig3}, for each channel of the input encoding feature of the fuzzy learning module, M fuzzy membership functions are applied to each feature point in the feature map, in order to convert the feature map extracted by the encoder into a fuzzy value, and the fuzzy value is a floating point number. The fuzzy membership functions remain constant across each channel but may vary across different channels. Each fuzzy membership function assigns a fuzzy class label to each feature point in each channel, describing its similarity to other feature points. Gaussian functions are employed as the fuzzy membership functions at the fuzzy membership function layer:

\begin{equation}
\label{17}
\begin{aligned} & F_{x,y,z,k,c}^{\prime}=e^{-\frac{\left(F_{x,y,z,c}-\mu_{\mathbf{k},c}\right)^{2}}{\sigma_{\mathbf{k},c}^{2}}}\\  & x=1...D,y=1...H,z=1...W\end{aligned}
\end{equation}

\noindent where $F_{x,y,z,c}$ represents a feature point in channel $C$ of the feature map, and $\text{(x,y,z)}$ denotes the corresponding position of the feature point. $\mu_{\mathbf{k},c}$ and $\sigma_{\mathbf{k},c}$ are the mean and standard deviation of the $k$-th Gaussian membership function, $\mu_{\mathbf{k},c}$ and $\sigma_{\mathbf{k},c}$ are randomly initialized and learnable, and $F_{x,y,z,k,c}^{\prime}$ denotes the $k$-th output fuzzy class label of the feature point at position $\text{(x,y,z)}$ in channel $C$. During training over the dataset, the membership functions are tuned to capture significant linguistic terms from the feature maps. Subsequently, the fuzzy rule "AND" is applied to all fuzzy class labels of each feature point to obtain the final fuzzy feature information $F_{x,y,z,c}^{\prime}$ of the encoded feature:

\begin{equation}
\label{18}
F_{x,y,z,c}^{\prime}=\prod_{k=1}^{M}F_{x,y,z,k,c}^{\prime}
\end{equation}

Equation (\ref{18}) applies the fuzzy rule “AND” according to the output of each fuzzy membership function to obtain the fuzzy feature information $F_{x,y,z,c}^{\prime}$ of each feature point. Consequently, the fuzzy feature map $F\in \mathbb{R}^{C\times D\times H\times W}$ after the feature graph extracted by U-Net encoder is processed by fuzzy learning module can be obtained.

\begin{figure}[!ht]
\centering
\includegraphics[width=0.5\textwidth]{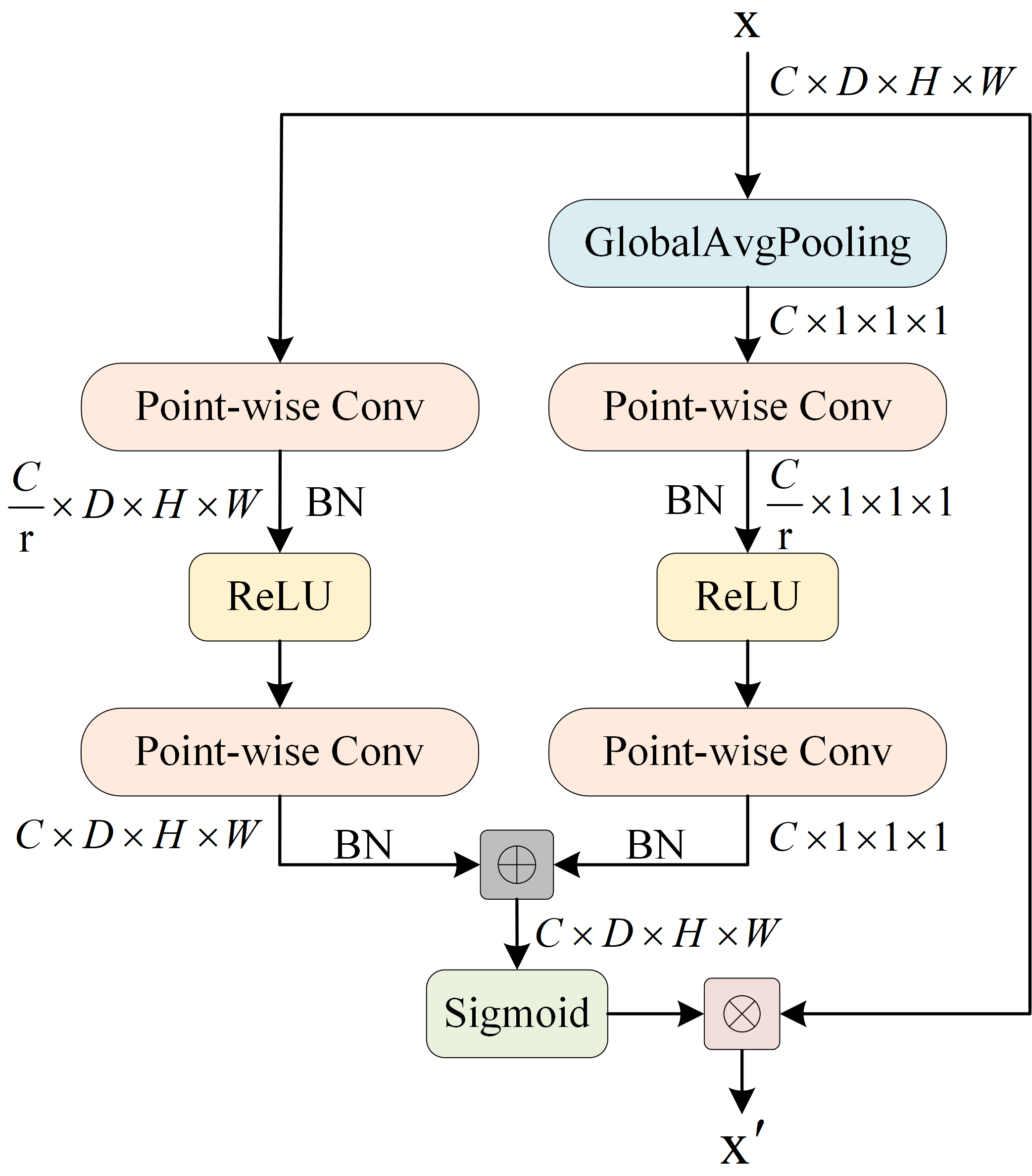}
\caption{Illustration of the proposed MS-CAM}
\label{fig4}
\end{figure}

As part of the network, the fuzzy learning module can be regarded as a fuzzy layer described by multiple parameters from Gaussian membership functions. It's worth noting that the parameters $\mu_{\mathbf{k},c}$ and $\sigma_{\mathbf{k},c}$ remain unchanged within the same channel but may vary across different channels. This is because the feature points in the same channel are extracted by the same convolutional kernel and share similar features, while those in different channels are obtained by using different convolution kernels and have different features. Additionally, to implement the fuzzy learning module, the number of fuzzy membership functions, denoted as $M$, is the only hyperparameter that needs to be preset.

Finally, as illustrated in the left of Figure \ref{fig3}, the output of the fuzzy learning module is a fuzzy tensor $F$, which shares the same size as the feature tensor $f_t$ processed by the U-Net. Consequently, they are fused through residual connections, as shown in Equation (\ref{19}).
\begin{equation}
\label{19}
F_t=BN(FLM(f_t))+BN(f_t)
\end{equation}

\noindent where the output feature map from the U-Net and the fuzzy tensor are processed through batch normalization (BN) to constrain their dynamic range. Subsequently, a simple addition operation is employed to integrate the fuzzy logic information.

By fuzzifying the encoded features with Gaussian membership functions, the module is able to handle noise and uncertainty in medical images using non-exclusive fuzzy membership values, which is more robust than deterministic features.

\begin{figure}[!ht]
\centering
\includegraphics[width=0.8\textwidth]{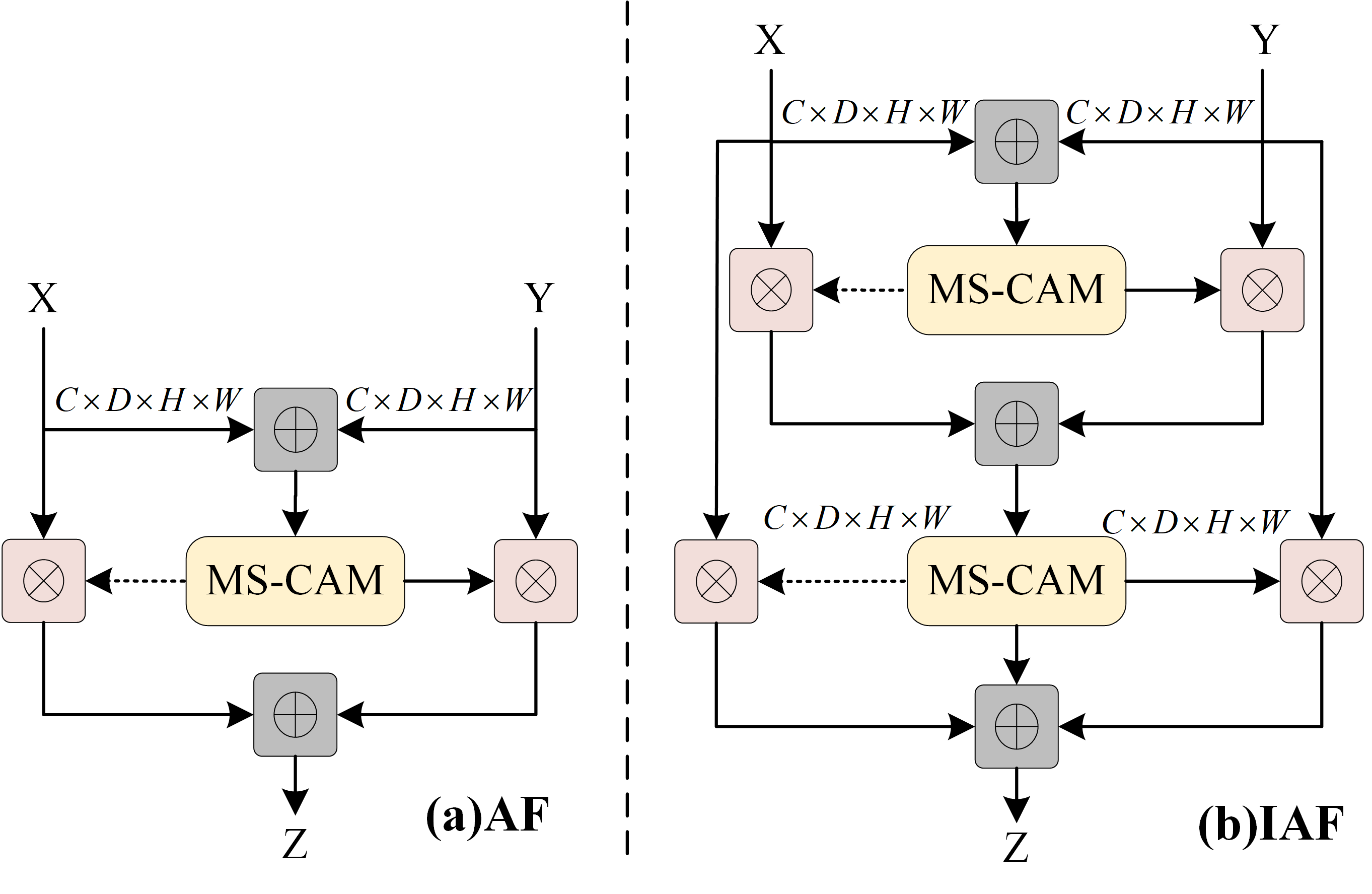}
\caption{AF and IAF architecture. (a) is the attention fusion structure. (b) is iterative attention fusion structure.}
\label{fig5}
\end{figure}

\subsection{Iterative attention fusion}
\label{3.3}
In traditional medical image segmentation tasks, the final prediction is used as the model's final segmentation result. In the reverse process of FDiff-Fusion, each time step generates a prediction result. However, the prediction results at each time step may differ in handling boundaries and details, and they may contain varying degrees of uncertainty and noise. Therefore, to improve the accuracy and robustness of the segmentation model, we integrate the segmentation maps obtained at each time step of the reverse process through an attention mechanism to obtain the final segmentation result. To this end, we designed a fusion module based on an iterative attention mechanism to optimize the model's prediction results. Traditional fusion methods typically employ simple addition or weighted fusion. In contrast, we introduce an iterative attention mechanism to achieve deeper fusion of each segmentation map obtained during the reverse process.

This section introduces iterative attention fusion from three parts. Firstly, it introduces multi-scale channel attention fusion(MS-CAM). Secondly, it introduces attention fusion(AF). Finally, iterative attention fusion(IAF) is proposed on the basis of multi-scale channel attention fusion and attention fusion.

\subsubsection{Multi-scale channel attention fusion}

The multi-scale channel attention module is divided into local attention and global attention branches. Its primary function is to enhance the representational capacity of input features through local and global feature weighting. Specifically, MS-CAM applies both local and global attention mechanisms to process input features, generating attention weights, and these weights are then used to weight the input features. As shown in Figure \ref{fig4}, in the local attention branch, given a feature $x\in\mathbb{R}^{C\times D\times H\times W}$ with $C$ channels, two layers of 1×1×1 convolutions (Conv3d) and Batch Normalization (BatchNorm3d), along with ReLU activation functions, are applied to process the input features. The local context $L(x)\in\mathbb{R}^{C\times D\times H\times W}$ can be calculated as follows:

\begin{equation}
\label{20}
L(x)=B(PWConv_2(\delta(B(PWConv_1(x)))))
\end{equation}
\noindent where $\delta$ represents the ReLU activation function and $B$ represents batch normalization. The core sizes of PWConv1 and PWConv2 are $\frac cr\times C\times1\times1$ and $C\times\frac{C}{r}\times1\times1$. This branch captures local spatial information, generating local features.

It is worth noting that $L(x)$ has the same shape as the input feature, preserving and emphasizing subtle details in low-level features. Similarly, in the global attention branch, the input feature $x$ is compressed to 1×1×1 in the spatial dimension through a global average pooling layer (GAP). Then, two layers of 1×1×1 convolutions, batch normalization, and ReLU activation function are used to extract the global context $G(x)\in\mathbb{R}^{C\times D\times H\times W}$:

\begin{equation}
\label{21}
G(x)=B(PWConv_2(\delta(B(PWConv_1(g(x))))))
\end{equation}

\noindent where $g(x)$ denotes the global average pooling (GAP). This branch captures global spatial information, generating global features. Given the global context $G(x)$ and the local context $L(x)$, the refined feature $x^{\prime}\in\mathbb{R}^{C\times D\times H\times W}$ of MS-CAM can be obtained from the following formula:

\begin{equation}
\label{22}
x^{\prime}=x\otimes\mathrm{M}(x)=x\otimes\sigma(\mathrm{L}(x)\oplus\mathrm{G}(x))
\end{equation}

\begin{algorithm}[H]
\caption{FDiff-Fusion\label{alg1}}
\KwIn{The medical image $I\in\mathbb{R}^{C\times D\times H\times W}$, the mask image $x_{0}\in\mathbb{R}^{C\times D\times H\times W}$}
\KwOut{The predicted segmentation label images $\hat{x}_{0}\in\mathbb{R}^{C\times D\times H\times W}$}
\For{epoch in max\_epoch}{
    {$\mathrm{t}\sim\mathrm{\emph{Uniform}}(\{1,...,\mathrm{T}\})$}\newline
    $\mathrm{\emph{Sampling~}}\beta_t$\newline
    {$\alpha_{t}=1-\beta_{t},\bar{\alpha}_{t}=\prod_{i=1}^{T}\alpha_{t}$}\newline
    {$\varepsilon_{t}\sim\mathcal{N}(0,1)$}\newline
    {$x_{t}=\sqrt{\bar{\alpha}_{t}}x_{0}+\sqrt{1-\bar{\alpha}_{t}}\bar{\varepsilon}_{t}$}\newline
    \For{t=T:1}{
    $f_I=Con\nu(I)$\newline
    $f_t=UNet\ encoder(x_{t})$\newline
    $f_t=f_t+f_I$\newline
    $F_t=BN(FLM(f_t))+BN(f_t)$\newline
    $x_{t-1}=UNet\ decoder(F_t)$\newline
    }
    $L(x)=BN(PWConv_2(ReLU(BN(PWConv_1(x)))))$\newline
    $G(x)=BN(PWConv_2(ReLU(BN(PWConv_1(g(x))))))$\newline
    $\mathrm{M}(x)=Sigmoid(\mathrm{L}(x)\oplus\mathrm{G}(x))$\newline
    \For{t=T-1:1}{
    $x_t\oplus x_{t-1}=M\left(x_t+x_{t-1}\right)\otimes x_{t-1}+\left(1-M\left(x_t+x_{t-1}\right)\right)\otimes x_t$\newline
    $x_{t-1}=M\left(x_t\oplus x_{t-1}\right)\otimes x_{t-1}+\left(1-M\left(x_t\oplus x_{t-1}\right)\right)\otimes x_t$\newline
    }
    $return\  \hat{x}_{0}$
}
\end{algorithm}

\noindent where $M(x)\in\mathbb{R}^{C\times D\times H\times W}$ represents the attention weights generated by MS-CAM. $\sigma$ is the Sigmoid function, $\oplus$ denotes broadcasted addition and $\otimes$ denotes element-wise multiplication.

\subsubsection{Attention fusion}
\label{Attention fusion}
Attention Fusion employs a multi-scale channel attention mechanism to weight the input features and then performs element-wise addition of the weighted features to generate a fused feature tensor. This process allows the model to adaptively focus on the important information within the input features, thereby enhancing feature representation capability and overall model performance.

Based on the multi-scale channel attention module, attention fusion can be expressed as:

\begin{equation}
\label{23}
\mathrm{Z=M(X\oplus Y)\otimes Y+\left(1-M(X\oplus Y)\right)\otimes X}
\end{equation}

\noindent where $Z\in\mathbb{R}^{C\times D\times H\times W}$ represents the fused features, $\oplus$ denotes element-wise summation of features, and $\otimes $ denotes element-wise multiplication. As shown in Figure \ref{fig5} (a), AF first sums the two input tensors element-wise and inputs them into the MS-CAM, where the output of the MS-CAM module is the attention weight ${M(X\oplus Y)}$, with the dashed line representing $1-{M(X\oplus Y)}$. Subsequently, the two input tensors are each multiplied element-wise with their corresponding attention weights to obtain the attention-weighted outputs. Finally, the two attention-weighted outputs are summed element-wise to generate the final fused feature tensor $Z$. Here, the range of the attention weights is between 0 and 1, and so is the range of $1-{M(X\oplus Y)}$, allowing the network to perform soft selection or weighted averaging between the inputs.

\subsubsection{Iterative attention fusion}
\label{Iterative attention fusion}
For the input features $X$ and $Y$, performing a simple element-wise summation of the input features may affect the final fusion weights. Since this is still a feature fusion problem, another attention module is used to fuse the input features. First, attention fusion is applied to the two input tensors to obtain an initial fused tensor $X\oplus Y$. Then, the initial fused tensor is input into the MS-CAM module for local and global attention processing to generate the attention weight ${M(X\oplus Y)}$. Finally, the two input tensors are multiplied element-wise with the attention weight, and the element-wise multiplication results are summed element-wise to obtain the final fused tensor $Z$. This two-stage method is referred to as iterative attention fusion, as shown in Figure \ref{fig5}(b). Iterative attention fusion can be represented as follows:

\begin{equation}
\label{24}
\mathrm{X\oplus Y=M(X+Y)\otimes Y+(1-M(X+Y))\otimes X}
\end{equation}

\begin{equation}
\label{24-25}
\mathrm{Z=M(X\oplus Y)\otimes Y+\left(1-M(X\oplus Y)\right)\otimes X}
\end{equation}

The complete process of FDiff-Fusion is shown in Algorithm 1.

\section{Experiment and result analysis\label{result}}
\subsection{Experimental setup}
We used two publicly available segmented datasets, including the BRATS 2020 Brain tumor dataset and the BTCV Abdominal Multi-organ dataset.

The BRATS 2020 Brain Tumor dataset is a publicly available dataset used for brain tumor segmentation \cite{r43,r44}, commonly employed in medical image processing. This dataset comprises 369 aligned multi-modal brain MRI scans from various medical centers, with each modality having a volume of 155×240×240. It includes four modalities: T1-weighted (T1), T1-weighted with contrast enhancement (T1Gd), T2-weighted (T2), and Fluid Attenuated Inversion Recovery (FLAIR) sequences. The images for each modality undergo registration and preprocessing to enable comparison at the same spatial locations. The BRATS 2020 Brain Tumor dataset also provides segmentation labels for brain tumors, including four types of tumor tissues: necrotic areas of glioblastoma, cysts, enhancing regions and normal tissues. The segmentation ratio for training, validation and testing sets is 0.7, 0.1 and 0.2, respectively.

The BTCV abdominal multi-organ dataset consists of 30 instances of 3D abdominal multi-organ images, with each 3D image containing segmentation targets for 13 organs \cite{r45}. All data are resampled to the same spatial resolution. Eighteen of the 3D abdominal multi-organ images are utilized for training the model, while the remaining twelve images are reserved for testing.

\subsection{Experimental parameter setting}
This experiment uses a workstation based on Windows11 system, the experiment platform is PC(Intel(R) Core(TM) i9-10940X@3.30GHz), the graphics card is NVIDIA GeForce RTX4090, and the memory capacity is 64G. The development tool is JetBrains PyCharm 2021.2.3 Professional version, using Python language to implement the relevant algorithms in the experiment.

During the initial stage of the experiment, data augmentation techniques such as random cropping, spatial padding, random flipping, intensity scaling and shifting were applied to all experimental data samples.

During the training phase, the loss function integrated Dice loss, BCE loss and MSE loss. For the BRATS 2020 brain tumor dataset, an AdamW optimizer with weight decay of 1e-3 was utilized, with a maximum iteration of 300, a batch size of 4, a learning rate of 1e-4, and cosine annealing used to update the learning rate. Random patches (each with a size of 96×96×96) were sampled for training in each iteration. During testing, DDIM sampling steps were set to 10, with each sample size being 96×96×96. The sliding window overlap rate was set to 0.5. For the BTCV abdominal multi-organ dataset, we adopts the AdamW optimizer with weight attenuation of 1e-3, the maximum number of iterations is 3000, the training batch size is 1, the learning rate is 2e-4, and the cosine annealing algorithm is used to update the learning rate. Each iteration randomly samples n patches (the size of each patch is 96×96×96) for training. In the test, the number of DDIM sampling steps is set to 10 and the size of each sample is 96×96×96. The sliding window overlap rate is 0.5.

\subsection{Experimental Evaluation Metrics}
To evaluate the segmentation performance of FDiff-Fusion, We used dice similarity coefficient (DSC), hausdorff distance at 95th percentile (HD95), the Jaccard index and recall as evaluation metrics.

The Dice similarity coefficient is a commonly used evaluation metric to measure the similarity between two sets. Its coefficient ranges from 0 to 1, where 1 indicates complete overlap, meaning the model's predicted segmentation result is identical to the ground truth segmentation result, and 0 indicates no overlap at all, meaning there is no common pixel between the two segmentation results. Typically, a Dice coefficient closer to 1 indicates higher segmentation quality, indicating that the model's predicted segmentation result is more similar to the ground truth segmentation result. Conversely, a Dice coefficient closer to 0 indicates lower segmentation quality, indicating greater disparity between the model's predicted segmentation result and the ground truth segmentation result. The Dice coefficient calculation formula is as follows:

\begin{equation}
\label{25}
DSC(A,B)=\frac{2|A\cap B|}{|A|+|B|}
\end{equation}

\noindent where $|A|$ and $|B|$ represent the ground truth labels and the model's predicted results, respectively. ${|A \cap B|}$ denotes the intersection of the model's predicted segmentation result and the ground truth segmentation result, while $|A|$ and $|B|$ represent the number of pixels in the two segmentation results, respectively.

In image segmentation tasks, the Hausdorff distance is a measure of similarity between two sets, considering the maximum distance between them. One set represents the ground truth segmentation result, while the other set represents the model's predicted segmentation result. The Hausdorff distance quantifies the maximum inconsistency between the two segmentation results, which is the distance from a point in one segmentation result to the farthest point in the other segmentation result.

HD95 is a percentile variant of the Hausdorff distance, which calculates the distance at the 95th percentile. In other words, HD95 represents the distance between the ground truth segmentation boundary and the model's predicted boundary, where 95\% of the distance values are less than or equal to the HD95 value. Therefore, a smaller HD95 indicates a better match between the model's predicted segmentation result and the ground truth boundary, indicating higher segmentation quality. The formula for HD95 calculation is as follows:

\begin{equation}
\label{26}
\mathrm{HD}(A,\mathrm{B})=\max\{\max_{g^{\prime}\in\mathrm{A},p^{\prime}\in\mathrm{B}}\|g^{\prime}-p^{\prime}\|,\max_{p^{\prime}\in\mathrm{A},g^{\prime}\in\mathrm{B}}\|p^{\prime}-g^{\prime}\|\}
\end{equation}

\noindent where $A$ and $B$ represent the ground truth label and the model's predicted result, respectively. $\|g'-p'\|$ and $\|p'-g'\|$ denote the distance between sets $A$ and $B$. HD95 utilizes the 95th percentile of distances between the ground truth and predicted surface point sets. Therefore, when calculating HD, the influence of a small fraction of outliers is minimized.

The Jaccard index is a statistical measure used to assess the similarity between two sample sets. It is defined as the ratio of the size of the intersection of two sets to the size of their union. The Jaccard index quantifies the similarity between two sets by computing the ratio of the number of common elements to the total number of distinct elements across both sets. The Jaccard coefficient can be expressed by the following formula:

\begin{equation}
\label{28}
J(A,B)=\frac{|A\cap B|}{|A\cup B|}
\end{equation}

\noindent where $A$ and $B$ represent the ground truth label and the model's predicted result, $|A\cap B|$ represents the size of the intersection of sets $A$ and $B$, $|A\cup B|$ represents the size of the union of sets $A$ and $B$.

Recall, also known as sensitivity, represents the proportion of actual positive samples that are correctly identified by the model. It focuses on measuring the model's ability to capture positive samples. The specific formula is as follows:The specific formula is as follows:

\begin{equation}
\label{29}
\mathrm{Recall}=\frac{TP}{TP+FN}
\end{equation}

\noindent where $TP$ (True Positives) represents the number of correctly predicted positive samples, and $FN$ (False Negatives) represents the number of actual positive samples that the model incorrectly predicted as negative.

\subsection{Parameter analysis}

\begin{table}
\caption{Segmentation performance for different values of parameters $t$, $epoch$, and $M$}
\label{table1}
\centering
\resizebox{0.5\linewidth}{!}{%
\begin{tabular}{ccccc} 
\hline
\multirow{2}{*}{} & \multicolumn{4}{c}{Average}                                       \\ 
\cline{2-5}
                  & DSC            & HD95          & Jaccard        & Recall          \\ 
\hline
t=800             & 66.69          & 9.31          & 61.47          & 58.83           \\
t=1000            & 83.82          & \textbf{6.60} & \textbf{75.88} & 83.27           \\
t=1500            & \textbf{84.19} & 7.44          & 75.16          & \textbf{86.67}  \\ 
\hline
epoch=2000        & 71.84          & 9.52          & 63.04          & 60.81           \\
epoch=3000        & 83.82          & \textbf{6.60} & \textbf{75.88} & 83.27           \\
epoch=4000        & \textbf{83.97} & 8.26          & 74.83          & \textbf{85.76}  \\ 
\hline
M=3               & 82.77          & 8.64          & 73.58          & 80.92           \\
M=5               & 83.82          & 6.60          & \textbf{75.88} & \textbf{83.27}  \\
M=7               & \textbf{83.94} & \textbf{5.90} & 75.64          & 82.86           \\
\hline
\end{tabular}
}
\end{table}

In FDiff-Fusion, several tunable parameters are involved, such as the number of time steps $t$, the number of epochs, the number of fuzzy membership functions $M$, and the hyperparameter sequence $\beta $ that controls the noise level. According to \cite{r46}, the hyperparameter sequence $\beta$ starts at 0.0001 and ends at 0.002. We primarily analyze the impact of varying $t$ , the number of $epochs$, and $M$ on the model's performance.

As shown in Table \ref{table1}, we conducted experiments with different values for the three parameters to compare their impact on the model's segmentation performance. From the table, it can be observed that when $t$ increases from 800 to 1000, the model's segmentation performance improves significantly, especially in terms of DSC and Jaccard index. Additionally, HD95 decreases substantially, indicating a significant increase in model accuracy. Meanwhile, the improvement in recall rate suggests that the model has a stronger ability to identify target regions. However, when $t$ is increased from 1000 to 1500, although there are improvements in DSC and recall rate, the Jaccard index slightly decreases and HD95 increases, indicating that the performance gains are not significant and may even be somewhat unstable.

Regarding the number of iterations, when the epochs increased from 2000 to 3000, the model's segmentation performance improved significantly. Both the DSC and Jaccard index saw substantial increases, HD95 decreased markedly, and the recall rate also improved significantly. However, when the epochs were further increased to 4000, the performance gains were not as evident, with some metrics even experiencing slight declines. Moreover, the increased number of epochs consumed a considerable amount of computational resources and time, indicating that increasing the number of iterations beyond a certain point has limited benefits for enhancing model performance.

Regarding the number of fuzzy membership functions, when $M$ increased from 3 to 5, there was a noticeable improvement in model performance as indicated by significant increases in DSC and Jaccard index, and a substantial decrease in HD95. However, when $M$ was increased from 5 to 7, although the segmentation performance of the model improved, the gains were only marginal, and this improvement was accompanied by a substantial consumption of computational resources and time.

Therefore, based on the above observations, we set the default values of $t$, $epoch$ and $M$ in the FDiff-Fusion model to 1000, 3000, and 5, respectively, to achieve a good balance between segmentation performance and computational time.

\begin{table}
\caption{Results of comparison of the BRATS 2020 brain tumor dataset}
\label{table2}
\centering
\resizebox{\linewidth}{!}{%
\begin{tabular}{ccccccccccc} 
\hline
\multirow{2}{*}{Models} & \multicolumn{3}{c}{DSC(\%)}                      & \multicolumn{3}{c}{HD95(mm)}                     & \multicolumn{4}{c}{Average}                                        \\ 
\cmidrule(lr){2-4}\cmidrule(lr){5-7}\cmidrule(lr){8-11}
                        & WT             & TC             & ET             & WT             & TC             & ET             & DSC            & HD95           & Jaccard        & Recall          \\ 
\hline
UNETR\cite{r11}                   & 88.94          & 79.89          & 72.24          & 4.315          & 5.952          & 4.729          & 80.36          & 4.999          & 72.74          & 86.37          \\
Swin UNETR\cite{r12}              & 89.71          & 80.21          & 73.25          & 2.974          & 4.859          & 4.696          & 81.05          & 4.177          & 73.61          & 88.50          \\
TransBTS\cite{r15}                & 89.36          & 81.24          & 75.19          & 3.660          & 3.094          & 3.604          & 81.93          & 3.453          & 73.59          & 87.17           \\
TransUNet\cite{r17}                & 89.21          & 82.47          & 75.82          & 3.146          & \textbf{2.891} & 3.621          & 82.50          & 3.219          & 74.98          & 89.92           \\
Diff-UNet\cite{r25}              & 89.37          & 83.15          & 76.92          & 2.118          & 3.489          & 2.271          & 83.14          & 2.626          & 75.04          & 89.34           \\
CKD-TransBTS\cite{r16}            & 89.83          & 84.05          & 76.96          & 2.419          & 3.447          & 3.018          & 83.61          & 2.961          & 75.61          & 89.67           \\
Slim UNETR\cite{r14}              & 90.17          & 83.87          & 77.43          & 2.207          & 3.311          & 2.714          & 83.82          & 2.744          & 76.05          & 90.81           \\
FDiff-Fusion            & \textbf{90.51} & \textbf{84.37} & \textbf{77.61} & \textbf{2.115} & 3.086          & \textbf{2.220} & \textbf{84.16} & \textbf{2.473} & \textbf{76.83} & \textbf{91.77}  \\
\hline
\end{tabular}
}
\end{table}

\begin{table}[htbp]\Huge
\caption{Results of comparison of BTCV abdominal multi-organ dataset}
\label{table3}
\centering
\resizebox{\linewidth}{!}{%
\begin{tabular}{ccccccccccccc} 
\hline
\multirow{2}{*}{Models} & \multicolumn{4}{c}{Average}                                      & \multicolumn{8}{c}{DSC(\%)}                                                                                                            \\ 
\cmidrule(lr){2-5}\cmidrule(lr){6-13}
                        & DSC            & HD95          & Jaccard        & Recall         & Spl         & Kid(L)      & Kid(R)      & Gall    & Liv          & Sto       & Aor        & Pan        \\ 
\hline
Vnet\cite{r47}                    & 68.81          & 27.45         & 61.27          & 69.17          & 80.56          & 77.1           & 80.75          & 51.87          & 87.84          & 56.98          & 75.34          & 40.05           \\
UNETR\cite{r11}                    & 78.42          & 23.66         & 72.74          & 78.35          & 86.03          & 82.97          & 77.35          & 72.61          & 94.58          & 74.64          & 85.32          & 53.91           \\
Swin UNETR\cite{r12}               & 79.64          & 18.91         & 72.95          & 78.86          & 86.76          & 83.51          & 78.79          & 72.62          & 94.19          & 77.62          & 86.03          & 57.61           \\
TransUNet\cite{r17}                & 78.03          & 30.72         & 72.81          & 79.01          & 85.62          & 82.64          & 77.98          & 64.54          & 93.94          & 73.61          & 87.21          & 55.34           \\
Diff-UNet\cite{r25}               & 82.17          & 10.23         & 74.35          & 81.74          & 86.61          & 83.19          & 83.79          & \textbf{75.32} & 94.65          & 75.02          & 86.62          & \textbf{72.16}  \\
CKD-TransBTS\cite{r16}            & 82.23          & 9.46          & 74.39          & 82.16          & 87.96          & 82.87          & 83.79          & 74.31          & 93.89          & 79.07          & 87.51          & 68.42           \\
Slim UNETR\cite{r14}              & 83.16          & 8.39          & 75.62          & 82.92          & 88.63          & 83.73          & 84.27          & 74.62          & 94.74          & 82.64          & 87.83          & 68.83           \\
FDiff-Fusion            & \textbf{83.82} & \textbf{6.60} & \textbf{75.88} & \textbf{83.27} & \textbf{89.08} & \textbf{84.12} & \textbf{84.51} & 74.01          & \textbf{95.33} & \textbf{84.02} & \textbf{88.48} & 70.98           \\
\hline
\end{tabular}
}
\end{table}

\subsection{Experimental results and analysis on different datasets}
To validate the effectiveness of FDiff-Fusion in medical image segmentation, we compared FDiff-Fusion with state-of-the-art (SOTA) segmentation models on the BRATS 2020 brain tumor dataset and the BTCV abdominal multi-organ dataset. The quantitative results are shown in Tables \ref{table2} and \ref{table3}. For the BRATS 2020 brain tumor dataset, our model was compared with widely used and recognized segmentation models, including CNN-based methods like UNETR, Swin UNETR, and Slim UNETR, Transformer-based methods like TransBTS, TransUNet, CKD-TransBTS, and the diffusion-based method Diff-UNet. For the BTCV abdominal multi-organ dataset, the same training and testing datasets were utilized, and comparisons were made with state-of-the-art models, including V-Net, UNETR, Swin UNETR, TransUNet, Diff-UNet, CKD-TransBTS, and Slim UNETR. All baseline models and FDiff-Fusion in this experiment were trained on the same dataset and the same computer hardware, and the best-performing models were selected based on the validation set. All quantitative and qualitative results are direct outputs from the models without any post-processing.

Table \ref{table2} presents the Dice scores and HD95 distances for the three sub-regions (whole tumor, tumor core, and enhancing tumor) on the BRATS 2020 brain tumor dataset, as well as the average Dice scores, average HD95 distances, average Jaccard indices and average Recall values. It is evident that our proposed method not only surpasses the SOTA methods in terms of the average values of these four evaluation metrics but also demonstrates superior performance in Dice scores and HD95 distances across the three sub-regions. The average Dice score across the three sub-regions reaches 84.16\%, representing an improvement of 0.34\% over the second-best method and 3.8\% over UNETR. Additionally, our model achieves the best results in HD95 distance, Jaccard index, and Recall, with values of 2.473, 76.83, and 91.77, respectively, clearly outperforming the state-of-the-art methods.

\begin{figure}[!ht]
\centering
\includegraphics[width=\textwidth]{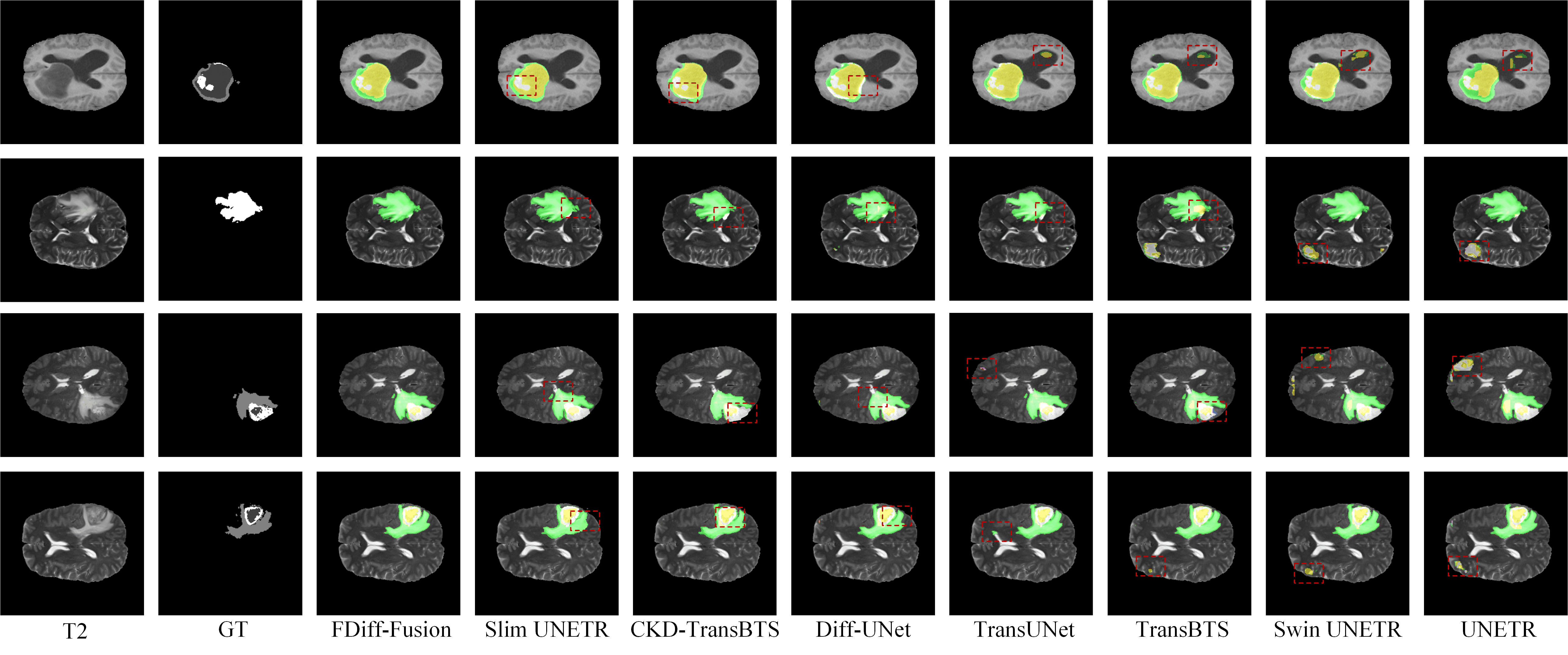}
\caption{Visualization of quantitative comparison of SOTA methods on the BRATS 2020 brain tumor dataset.}
\label{fig6}
\end{figure}

Table \ref{table3} displays the Dice scores for eight abdominal organs(Spleen, Kidney(L), Kidney(R), Gallbladder, Liver, Stomach, Aorta, Pancreas), as well as the average Dice score, average HD95 distance, average Jaccard index, and average Recall, comparing FDiff-Fusion with state-of-the-art segmentation methods. As shown in Table \ref{table3}, it is evident that FDiff-Fusion achieves the best average Dice score, average HD95 distance, average Jaccard index, and average Recall across these eight organs. Specifically, Our method ranks first in Dice score for 6 organs, with the average Dice score, average HD95 distance, average Jaccard index, and average Recall reaching 83.82\%, 6.60, 75.88, and 83.27 respectively, indicating the superiority of our method in the BTCV abdominal multi-organ dataset compared to the state-of-the-art methods. Although FDiff-Fusion does not rank first in Dice scores for the Pancreas and Gallbladder, with scores of 70.98\% and 74.01\% respectively, these scores are only slightly lower than the best scores of 72.16\% for the Pancreas and 75.32\% for the Gallbladder.

\begin{figure}[!ht]
\centering
\includegraphics[width=\textwidth]{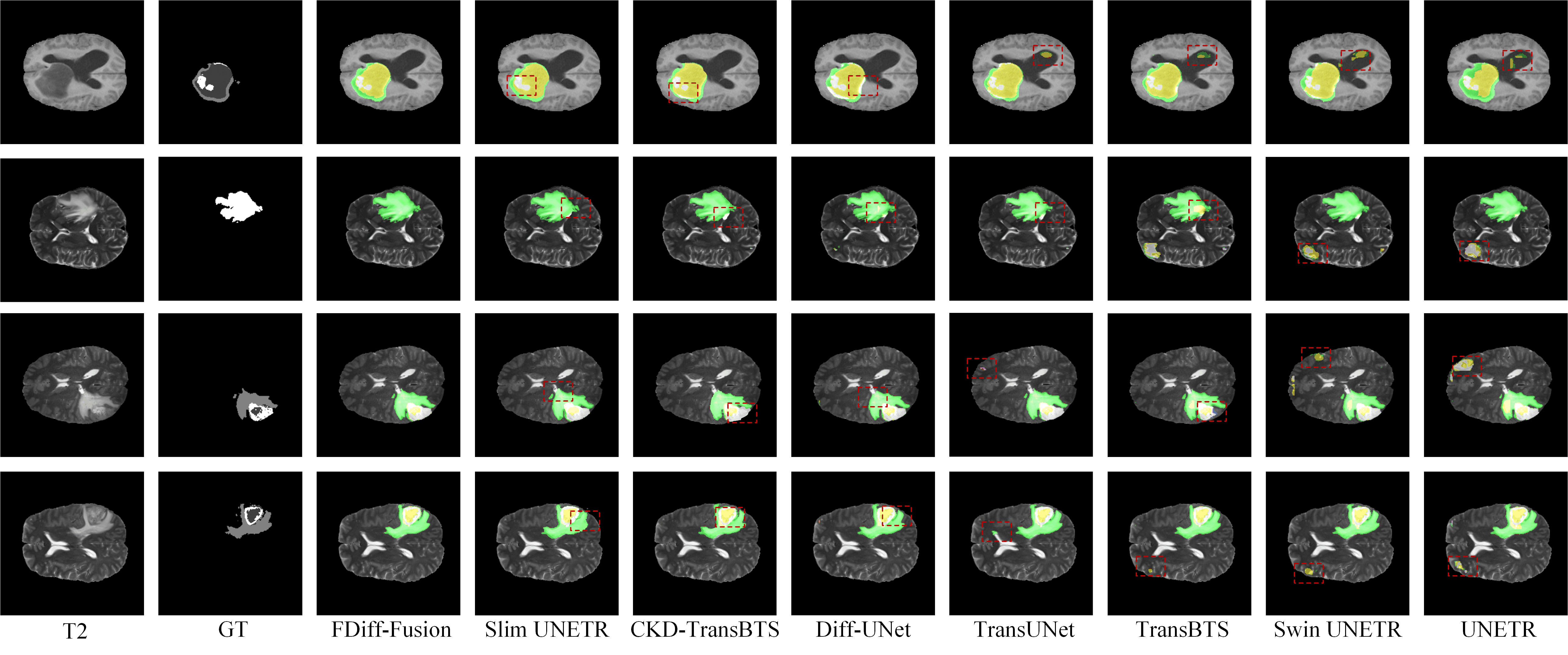}
\caption{Visualization of quantitative comparison of SOTA methods on the BTCV abdominal multi-organ dataset.}
\label{fig6-7}
\end{figure}

Figure \ref{fig6} illustrates the qualitative comparison results between FDiff-Fusion and several baseline models on the BRATS 2020 brain tumor dataset. Here, we use T2 as the segmentation background, and GT represents the ground truth label. In the first row, our proposed model successfully suppresses false positive results compared to other baseline models (marked by red dashed boxes). In the second row, compared to other SOTA models, FDiff-Fusion generates more precise segmentation results with less noise and segmentation boundaries closer to the ground truth label, thanks to the design of the fuzzy learning module. In the third row, our model achieves more accurate and complete segmentation boundaries, attributed to the proposed iterative attention fusion. It can be seen from the fourth row that the segmentation result of our model is closer to the ground turth label. The experiments demonstrate that the FDiff-Fusion model achieves more precise segmentation results on segmentation boundaries and subtle segmentation targets, while baseline models may miss some segmentation target areas or include other non-segmentation target areas in the segmentation results. Moreover, our proposed model also exhibits significant superiority in the precise identification and segmentation of target area boundaries.

In order to better demonstrate the segmentation performance of our model for the BTCV abdominal multi-organ dataset, we selected a slice of the segmentation graph for visualization. It is clear from Figure \ref{fig6-7} that FDiff-Fusion performs well on the BTCV abdominal multi-organ dataset. Compared with the baseline model, its segmentation results are more precise and accurate, and more consistent with the ground truth label. This advantage comes from the fuzzy learning module and the iterative attention fusion module designed by us. In the input image, it is able to capture key morphological and structural information and fuse this information into the segmentation results, resulting in finer segmentation boundaries and more accurate organ regions.

\subsection{Ablation study}
We conducted comprehensive ablation studies to validate the effectiveness of the proposed fuzzy learning module and iterative attention fusion. The results are shown in Table \ref{table4}, where basic represents a baseline model without the addition of fuzzy learning and attention fusion, FLM represents the fuzzy learning module, AF represents attention fusion and IAF represents iterative attention fusion. It can be observed that the average Dice score of basic+FLM is 0.65\% higher than that of basic, and on average HD95 distance, it is 0.822 lower than basic. This indicates that the designed FLM in this paper effectively addresses the uncertainty of segmentation boundaries and region fuzziness in medical images, thereby significantly improving the segmentation accuracy of the model. Furthermore, there is a noticeable improvement in both the average Dice score and average HD95 distance for basic+FLM+AF compared to basic+FLM. Moreover, FDiff-Fusion exhibits better performance in both average Dice score and average HD95 distance compared to basic+FLM+AF, highlighting the significant effectiveness of the introduced attention fusion mechanism in enhancing the segmentation performance of the model. From Table 3, it can be observed that the average Dice score of FDiff-Fusion is 1.24\% higher than basic, and the average HD95 distance is 1.518 lower than basic, demonstrating the superior effectiveness of the proposed fuzzy learning module and iterative attention mechanism for medical image segmentation.

\begin{table}[htbp]\scriptsize
\caption{Ablation studies of different modules on the BraTS 2020 brain tumor dataset}
\label{table4}
\centering
\resizebox{\linewidth}{!}{%
\begin{tabular}{ccccccccc} 
\midrule
\multirow{2}{*}{Module} & \multicolumn{4}{c}{DSC(\%)}                                       & \multicolumn{4}{c}{HD95(mm)}                                       \\ 
\cmidrule(lr){2-5}\cmidrule(lr){6-9}
                        & WT             & TC             & ET             & Average        & WT             & TC             & ET             & Average         \\ 
\midrule
basic                   & 89.24          & 82.98          & 76.54          & 82.92          & 3.754          & 3.841          & 4.379          & 3.991           \\
basic+FLM               & 90.12          & 83.64          & 76.98          & 83.57          & 3.008          & 3.294          & 3.206          & 3.169           \\
basic+FLM+AF            & 90.42          & 83.88          & 77.46          & 83.92          & 2.239          & \textbf{3.071} & 2.716          & 2.675           \\
FDiff-Fusion            & \textbf{90.51} & \textbf{84.37} & \textbf{77.61} & \textbf{84.16} & \textbf{2.115} & 3.086          & \textbf{2.220} & \textbf{2.473}  \\
\midrule
\end{tabular}
}
\end{table}

\begin{figure}[!ht]
\centering
\includegraphics[width=\textwidth]{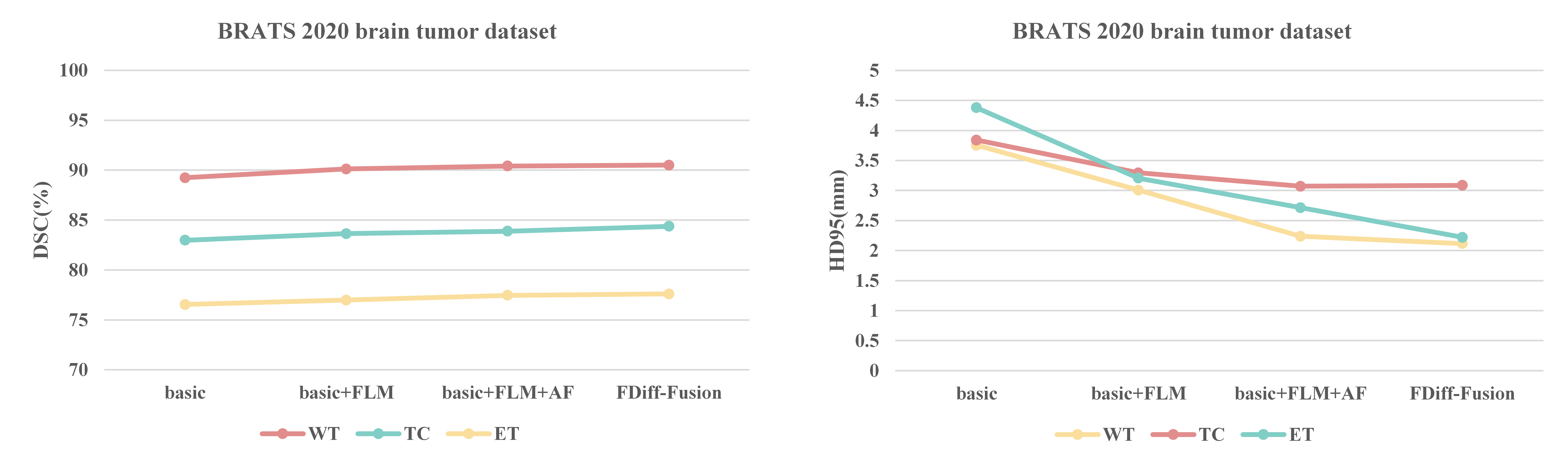}
\caption{Line chart comparison of FDiff-Fusion ablation studies on the BRATS 2020 brain tumor dataset.}
\label{fig8}
\end{figure}

We also visually presented the results of the ablation study using line plots, as shown in Figure \ref{fig8}. The left chart depicts the line plot for Dice scores, while the right chart shows the line plot for HD95 distances. By observing the variations of different segmentation targets in the line plots, it can be seen that the segmentation performance of the FDiff-Fusion model achieves the best results.

\begin{figure}[!ht]
\centering
\includegraphics[width=\textwidth]{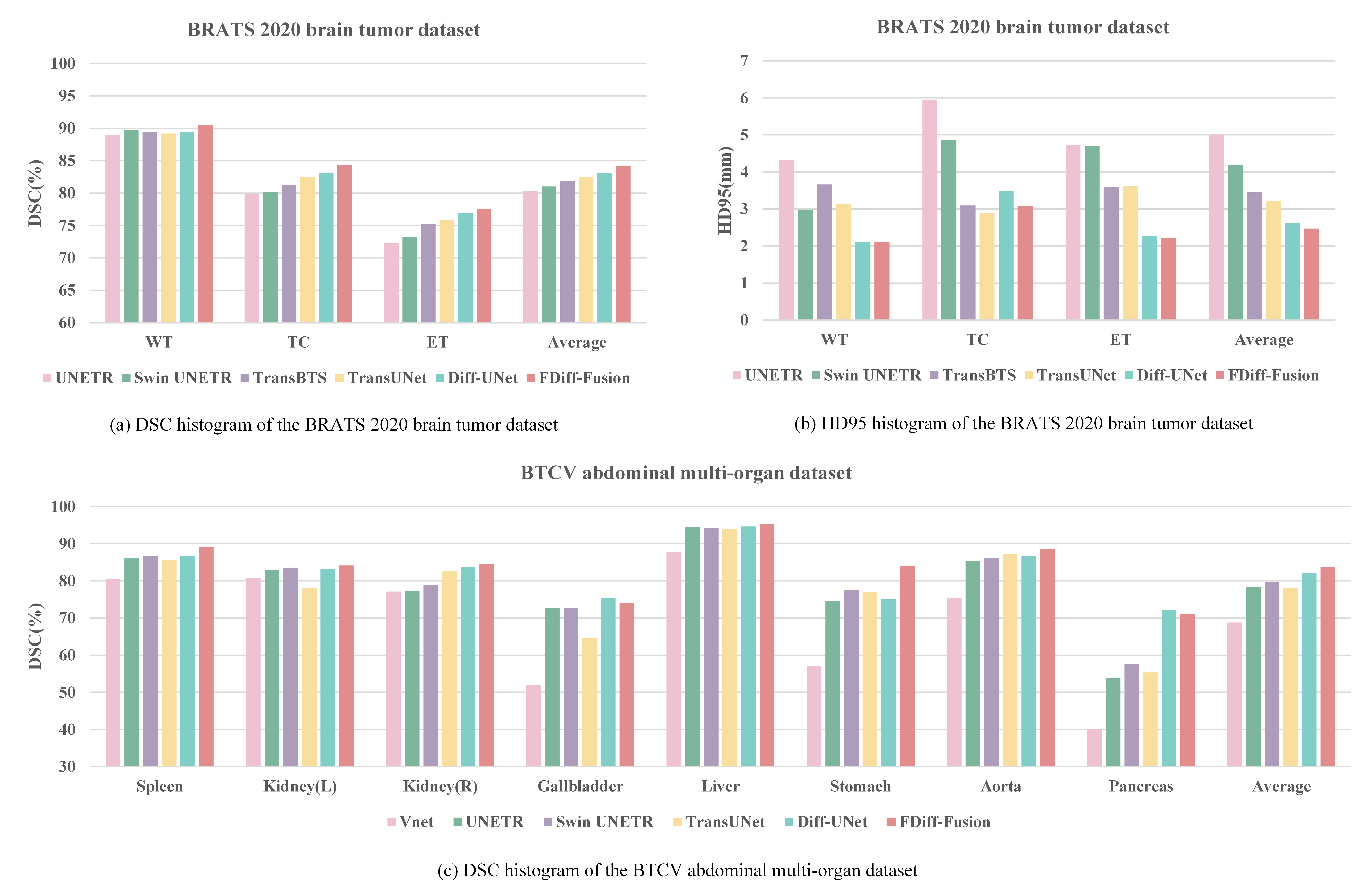}
\caption{Histogram comparison of segmentation results of different models on different data sets.}
\label{fig7}
\end{figure}

\subsection{Computational efficiency analysis}
Firstly, we acknowledge that the denoising diffusion model and the fuzzy learning module may impact the training time of the model. However, we have implemented various optimization strategies to mitigate this effect, such as employing caching mechanisms to reduce data reading time.

Compared to baseline models, our method may require longer training times due to the introduction of more complex computations and additional parameters aimed at achieving higher accuracy. Nevertheless, this increased computational cost is justified, as our experimental results demonstrate significant improvements in segmentation performance.

Moreover, it is important to emphasize that the increase in training time does not directly affect the efficiency in practical applications. Training is a one-time process, whereas in practical applications, the inference phase speed is a more critical consideration. Our method maintains efficient performance during the inference phase, ensuring it remains fast and effective in real-world applications.

\subsection{Discussion}
To provide a clearer demonstration of FDiff-Fusion's superiority in 3D medical image segmentation, we evaluates the model using bar charts. As shown in Figure \ref{fig7}, panels (a) and (b) respectively represent bar charts of Dice scores and HD95 distances on the BRATS 2020 brain tumor dataset, while panel (c) depicts a bar chart of Dice scores on the BTCV abdominal multi-organ dataset. Based on the heights of the bars in the bar chart, it can be observed that the FDiff-Fusion model exhibits superior segmentation performance.

\begin{figure}[!ht]
\centering
\includegraphics[width=\textwidth]{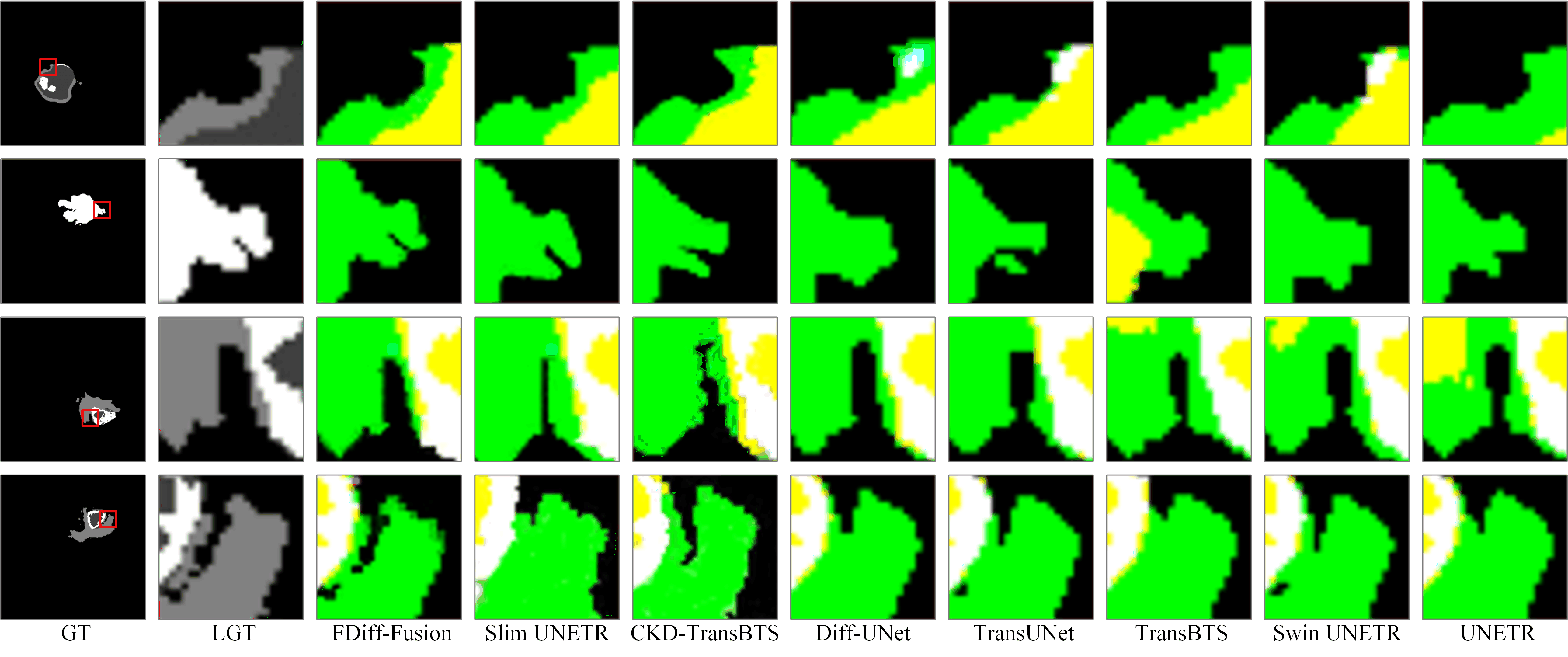}
\caption{Local visualization of segmentation results from the BRATS 2020 brain tumor dataset.}
\label{fig10}
\end{figure}

Furthermore, to vividly demonstrate the advantages of the FDiff-Fusion over other baseline models in segmenting medical images, we performed a detailed visualization of the local segmentation results of the BRATS 2020 brain tumor dataset. By selectively cropping portions of segmented images and highlighting minor differences in segmentation boundary areas, we were able to intuitively observe our model's ability to handle the common challenges of boundary uncertainty and fuzzy regions in medical images, as shown in Figure \ref{fig10}. In this image, GT represents the ground truth labels, while LGT denotes the image after local cropping of the ground truth labels. From this visualization result, we can clearly see that compared with other benchmark models, our proposed FDiff-Fusion model can more accurately classify the classification of each pixel in medical images. This observation not only reinforces the robustness and accuracy of our model but also highlights its superiority in addressing the complexities of medical image segmentation.

\begin{figure}[!ht]
\centering
\includegraphics[width=0.7\textwidth]{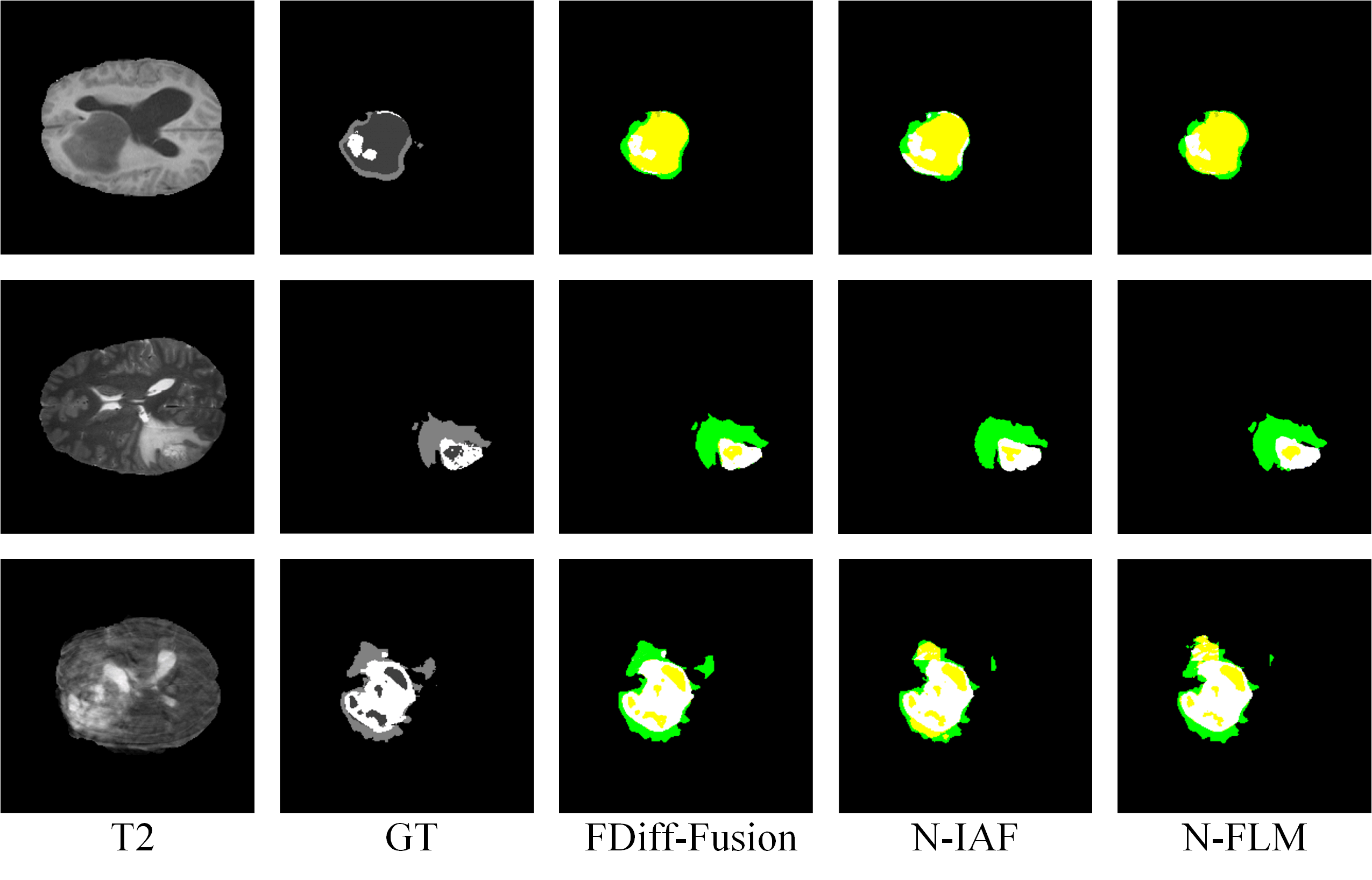}
\caption{Visualization of ablation experiment of FLM and IAF.}
\label{fig11}
\end{figure}

We also conducted visualization of the ablation experiments for the designed Iterative Attention Fusion module and the Fuzzy Learning Module . Specifically, we first removed the IAF module from the model and performed experiments. Through this ablation experiment, we were able to observe changes in model performance and the specific contribution of the IAF module to the overall segmentation. As shown in Figure \ref{fig11}, N-IAF denotes the removal of the IAF module, and N-FLM denotes the removal of the FLM. The experimental results indicate that removing the IAF module results in a significant drop in segmentation accuracy, demonstrating that this module plays a critical role in enhancing the model's robustness. Additionally, we removed the FLM and conducted experiments. By ablating the FLM, we could assess its practical utility in segmentation. The experimental results show that removing the FLM significantly weakens the model's ability to capture and focus on the edge features of the segmentation targets, leading to a decline in overall performance. This further verifies the importance of the FLM in the model.

\section{Conclusion and Outlook\label{conclusion}}
This paper presents a denoising diffusion fusion network based on fuzzy learning for 3D medical image segmentation, named FDiff-Fusion. To address the common phenomenon of segmentation boundary uncertainty and region fuzziness in medical images, we propose two novel techniques to ensure the performance of the model, namely fuzzy learning and iterative attention fusion. By adding fuzzy learning in the skip pathways of the U-Net network, multiple fuzzy membership functions are set for the input encoded features to describe the similarity between feature points. Fuzzy rules are then applied to the fuzzy membership functions to enhance the model's ability to model uncertain boundaries and fuzzy regions. Additionally, we design a method based on iterative attention feature fusion, which adds local contextual information to the global contextual information in the attention module to fuse the prediction results of each denoising time step, thereby improving the accuracy and robustness of the model segmentation. Experimental results on two different benchmark datasets demonstrate the superiority of our model over state-of-the-art models, providing an effective solution for the field of medical image segmentation.

While FDiff-Fusion has achieved significant success in the field of medical image segmentation, there are still many directions for further exploration and improvement. For example, our model, due to the introduction of complex denoising diffusion models, the fuzzy learning module, and iterative attention fusion, requires substantial computational resources during training. This may limit its applicability in environments with limited computational resources. Additionally, our model relies on high-quality annotated data during the training process, which means that its performance may be impacted in scenarios where data is scarce or of low quality. Future work can be extended to more types of medical image tasks, such as tumor segmentation, organ segmentation, and lesion detection, etc. In addition, we also believe that ensemble learning, semi-supervised and unsupervised learning can be used as potential directions for future research work to further improve the performance and applicability of our method.

\section*{Acknowledgments}
This work was supported in part by the National Natural Science Foundation of China under Grants 61976120 and 62102199, in part by the Natural Science Foundation of Jiangsu Provinceunder Grant BK20231337, in part by the Natural Science Key Foundation of Jiangsu Education Department under Grant 21KJA510004, in part by the China Postdoctoral Science Foundation under Grant 2022M711716 and in part by the Postgraduate Research \& PracticeInnovation Program of Jiangsu Province under Grant SJCX24\_2017

\bibliographystyle{elsarticle-num} 
\bibliography{ref}

\end{document}